\pdfoutput=1
\documentclass[11pt]{article}
\usepackage[preprint]{acl}
\usepackage{times}
\usepackage{latexsym}
\usepackage[T1]{fontenc}
\usepackage[utf8]{inputenc}
\usepackage{microtype}
\usepackage{inconsolata}
\usepackage{booktabs}
\usepackage{array}
\usepackage{graphicx}
\usepackage{amsmath}
\usepackage{xcolor}

\expandafter\def\expandafter\UrlBreaks\expandafter{\UrlBreaks\do\-}
\graphicspath{{figures/}}

\title{Temperature Fragility and the Conditional Benefits of Truncation Sampling}
\author{Francesco La Rosa \\ University of Edinburgh \\ \texttt{larosafrancesco289@gmail.com}}

\begin{document}
\maketitle

\begin{abstract}
Large language models generate text by sampling each token from a predicted distribution, and a
temperature parameter sets how far the draw strays from the most probable tokens. Truncation
samplers such as top-$p$ and min-$p$ discard the least probable tokens before the draw, so that
sampling at high temperature stays coherent. Their reported accuracy gains come from temperatures
of 1.5 to 3, while the defaults of deployed systems cluster between 0.6 and 1.0. Whether they change
accuracy at those defaults, and for which models, has not been measured. We test thirteen open-weight
models on GSM8K and MMLU-Pro at temperatures 0.7, 1.0, and 1.3 in one controlled pipeline, ten
of them under eight decoding configurations. Six of the thirteen models lose 17 to 38 accuracy
points on MMLU-Pro between 0.7 and 1.3, and the other seven lose at most 10. The lost accuracy
comes from generations that run to the token limit or never state an answer. These results suggest
that truncation samplers improve accuracy primarily when higher temperatures substantially degrade
model performance. Where accuracy remains stable across temperatures, none of the tested
truncation samplers improves on plain temperature sampling.
\end{abstract}

\section{Introduction}
\label{sec:intro}

Sampling is the default way deployed language models generate text \citep{shi2024decoding}.
At each step the model assigns a probability to every token in its vocabulary, and the decoder
draws one from that distribution. A temperature parameter reshapes the distribution before the
draw. Low temperatures concentrate probability on the few most likely tokens. High temperatures
spread it toward the tail of the distribution, the many tokens the model rates as unlikely, where
its estimates are least reliable \citep{holtzman2020curious,hewitt2022truncation}. The typical
defaults of model vendors and inference engines cluster between 0.6 and 1.0
\citep{renze2024temperature,hochlehnert2025sober},\footnote{\texttt{generation\_config.json}
of \texttt{meta-llama/Llama-3.1-8B-Instruct} sets 0.6; llama.cpp's \texttt{common/sampling.h}
sets 0.8; the OpenAI chat-completions and Anthropic messages API references set 1.0. All read
July 2026. Recommendations outside this range exist: Mistral's instruct model cards suggest
0.15.} and every engine implements truncation samplers beside them. Truncation samplers,
among them top-$p$, min-$p$, and top-$n\sigma$, discard the tail of the distribution before the
draw, each by its own criterion. They were designed to keep open-ended text coherent at high
temperature \citep{garcesarias2025decoding}, and they are now also evaluated on tasks with one
checkable answer, where the claim is accuracy: min-$p$ reports gains on GSM8K
\citep{nguyen2024minp}, and top-$n\sigma$ reports accuracy held up to a temperature of 3.0
\citep{tang2024topn}.

Both results rest on one model and one range of temperatures. Top-$n\sigma$ was evaluated on
Llama-3-8B \citep{tang2024topn} and min-$p$ on Mistral-7B \citep{nguyen2024minp}, at
temperatures of 1.5 to 3, above the defaults systems set. A reanalysis that searched the same
number of hyperparameter settings for every sampler found the gains of min-$p$ indistinguishable
from those of the other samplers \citep{schaeffer2025minp}. The question cannot be settled by
comparing across papers, because benchmark scores move with the inference backend even under
greedy decoding \citep{masoudian2026backend}, and with the prompt, the parser, and the item set.
Settling it requires one pipeline in which everything except the decoding configuration is held
fixed. It also requires enough models to show whether the answer depends on the model.

\begin{quote}
\textbf{Research question:} Under what model and temperature conditions do truncation samplers
improve accuracy over plain temperature sampling?
\end{quote}

We test thirteen open-weight models from the Llama-3, Mistral, Qwen, Gemma, and OLMo families
on GSM8K and MMLU-Pro at temperatures 0.7, 1.0, and 1.3, in one pipeline with fixed items,
prompts, token budgets, and parser. Ten of the models also run under eight decoding
configurations: greedy decoding, plain temperature sampling, and six truncation configurations.
Section~\ref{sec:design} gives the design. Two questions follow. The first is whether a model
keeps its accuracy between 0.7 and 1.3 under plain sampling, and what the lost accuracy looks
like when it does not. The second is whether any truncation sampler improves on plain temperature
sampling between 0.7 and 1.3.

We find that temperature sensitivity differs by model. Between 0.7 and 1.3, six of the thirteen
models lose 17 to 38 accuracy points on MMLU-Pro under plain sampling, and the other seven lose
at most 10. The loss takes the form of a collapse of the output. On the six fragile models, the
share of generations that run to the token limit or never state an answer rises by 26 to 78
points, while the share of generations that state a wrong answer does not rise. Hermes-3, a
different post-training of the same base model as Llama-3.1-8B, loses half as much accuracy as
Llama-3.1-8B does. Truncation gains follow the
collapse. On the models that hold their accuracy, no truncation sampler improves on plain
temperature sampling at 0.7 or 1.0. On the models that collapse, every truncation sampler
improves accuracy at 1.3, and at temperatures up to 2.0 truncation delays or prevents the
collapse. The gains reported for truncation samplers are therefore real, and they are recoveries
from a collapse that only some models suffer and that begins above the temperatures systems
use. This work makes three contributions:

\begin{itemize}
\item \textbf{A controlled comparison of truncation samplers at the temperatures systems use.}
On the robust models, no truncation sampler improves on plain temperature sampling at 0.7 or
1.0 on GSM8K and MMLU-Pro, and an upper bound states how large a gain the results leave
possible.
\item \textbf{Temperature fragility as a property of the model.} Six of the thirteen models
collapse between 1.0 and 1.3 into generations that run to the token limit or never state an
answer. The collapse replicates across inference engine and precision. A different
post-training of the same base model halves it. A plain temperature test identifies the
fragile models before any sampler is compared.
\item \textbf{The condition under which truncation samplers help.} On the collapsing models,
every truncation sampler improves accuracy at 1.3 and delays the collapse at temperatures up
to 2.0. This locates the gains reported for truncation samplers above the temperatures systems
use, which reconciles those reports with the absence of gains below.
\end{itemize}

\section{Related work}
\label{sec:background}

\textbf{Truncation samplers.} A truncation sampler removes the least probable tokens from the
distribution before each token is sampled. Top-$k$ keeps the $k$ most probable tokens
\citep{fan2018hierarchical}. Top-$p$, or nucleus sampling, keeps the smallest set of tokens
whose probabilities sum to $p$, so the set grows when the distribution is flat
\citep{holtzman2020curious}. Min-$p$ keeps every token whose probability is at least a fixed
fraction of the largest \citep{nguyen2024minp}. Top-$n\sigma$ works on the logits, the model's
scores before they are converted to probabilities, and keeps the tokens within $n$ standard
deviations of the largest logit \citep{tang2024topn}. Temperature rescales all logits equally,
so top-$n\sigma$ keeps the same tokens at every temperature, and so does top-$k$, because
scaling does not change the order of the tokens. Top-$p$ and min-$p$ decide from the
probabilities themselves, which temperature changes, so for them the order of scaling and
truncation matters. A sampler applied after temperature scaling sees the scaled distribution
and keeps a different set of tokens from the same sampler applied before it. Our main
configurations apply temperature first, and for top-$p$ and min-$p$ we also run the reverse
order, which is the default in llama.cpp.

\textbf{Temperature-invariant samplers.} The parameter of top-$p$ or min-$p$ fixes how much of
the distribution is kept, and temperature reshapes the distribution. The same value therefore
keeps a different set of tokens at each temperature, and a value that works well at one
temperature need not work well at another. Two samplers published in 2026 begin from this
observation. $p$-less sampling \citep{tan2026pless} and Min-$k$ \citep{ding2026mink} each
decide which tokens to keep from the shape of the distribution at every step, so that
temperature does not change which tokens are kept. Neither paper measures whether that matters
at the temperatures systems use. The evaluation of Min-$k$ begins at temperature 1.0, where
top-$k$, top-$p$, and min-$p$ score within a point of one another on GSM8K, and the samplers
separate only above temperature 2 \citep{ding2026mink}.

\textbf{Evaluations of decoding and temperature.} A survey of decoding methods reports that
the best method changes with the task and the model \citep{shi2024decoding}. The best configuration for
open-ended generation varies by model \citep{garcesarias2025decoding}, and so does the gap
between greedy decoding and sampling on reasoning benchmarks \citep{song2024goodbad}. Studies
of temperature either stop at 1.0 or use tasks with little accuracy to lose.
\citet{renze2024temperature} find no effect of temperature between 0.0 and 1.0 on
multiple-choice problems. \citet{grover2026reliability} concludes that instruction tuning is
the main determinant of robustness to temperature, on synthetic tasks where the most robust
models score 11 to 30\% and with no truncation sampler tested. \citet{fastowski2025confidence}
link the entropy of the token distribution to temperature sensitivity on factual questions,
and \citet{troshin2025temperature} attribute high-temperature losses on mathematics to wrong
tokens at a few critical positions.

\textbf{Conditions of measurement.} One recent study shapes the design of our comparison.
\citet{masoudian2026backend} run three instruction-tuned models through five inference
frameworks on six benchmarks and find that the framework changes the scores even under greedy
decoding, by amounts that depend on the model. A comparison across models and samplers must
therefore hold the framework fixed, or the differences it reports may belong to the framework.
We run every configuration through one engine, llama.cpp \citep{gerganov2023llamacpp}. To check whether the collapse we
observe comes from the engine, we rerun the most fragile model through a second engine at full
precision. Our models also run quantized, with weights stored at reduced precision to fit in less memory.
\citet{prasad2026quantization} find that quantization
does not change how much a model loses to temperature, and our grid at four quantization
levels agrees (Appendix~\ref{app:temp}).

\section{Experimental design}
\label{sec:design}

Table~\ref{tab:design} lists the six runs. All of them use the same items, prompts, token
budgets, and parser, so a difference in accuracy between two runs comes from the factor that
differs between them, up to the sampling variation that the confidence intervals quantify.

\begin{table*}[t]\centering
\caption{The six runs, 232,700 graded generations in total. Every run uses both tasks with the
same 50 fixed items each and three repetitions of every sampled configuration (one for greedy
decoding), except the stratified subset, which uses 50 fresh MMLU-Pro items. ``8''
configurations are the eight described under Decoding. The main grid also includes a
16-bit GGUF run of Qwen3-4B beside its four quantized runs.}
\label{tab:design}
\footnotesize\setlength{\tabcolsep}{3pt}\resizebox{\textwidth}{!}{\begin{tabular}{@{}llllrr@{}}
\toprule
Run & Models & Precision and engine & Configurations & $T$ & Generations \\
\midrule
Main grid & 7, from 5 families & Q8, Q6, Q4, Q3; llama.cpp & 8 & 0.7, 1.0, 1.3 & 185,600 \\
Temperature test & 6 more & Q8; llama.cpp & greedy, plain temp. & 0.7, 1.0, 1.3 & 6,000 \\
Flagged-model grid & 3 of the 6 & Q8; llama.cpp & 8 & 0.7, 1.0, 1.3 & 19,200 \\
Engine check & Llama-3.2, Llama-3.1, Qwen3-4B & BF16, INT8 (Transformers); F16, Q8 (llama.cpp) & greedy, plain temp. & 0.7, 1.0, 1.3 & 6,000 \\
Temperature sweep & Llama-3.1, Qwen2.5, Gemma-3 & Q8; llama.cpp & 4 & 1.3, 1.5, 1.7, 2.0 & 14,400 \\
Stratified subset & Llama-3.1, Qwen2.5, Gemma-3 & Q8; llama.cpp & greedy, plain temp. & 0.7, 1.0, 1.3 & 1,500 \\
\bottomrule
\end{tabular}
}
\end{table*}

\textbf{Decoding.} Every model in the main grid runs under eight decoding configurations. Two
are baselines, greedy decoding and plain temperature sampling with no truncation. Four apply
one truncation sampler after temperature scaling, each at the setting its source paper uses:
top-$p$ 0.95 \citep{holtzman2020curious}, top-$k$ 40 \citep{radford2019gpt2}, min-$p$ 0.05
\citep{nguyen2024minp}, and top-$n\sigma$ 1.0 \citep{tang2024topn}. The last two repeat top-$p$
and min-$p$ with truncation applied before temperature scaling, the default order in llama.cpp,
so both orders are tested for the two samplers where the order matters
(Section~\ref{sec:background}). Temperatures 0.7 and 1.0 bracket the defaults of
Section~\ref{sec:intro}, and 1.3 sits above them, to show what happens when a model is run past
its default range.

\textbf{Tasks.} GSM8K is a set of grade-school word problems with a numeric answer
\citep{cobbe2021gsm8k}. We use the zero-shot chain-of-thought prompt of Meta's Llama evaluation
suite, which asks for reasoning without worked examples \citep{meta2024llamaevals}. MMLU-Pro is
ten-option multiple choice across 14 categories \citep{wang2024mmlupro}, prompted the same way.
Both tasks need a chain of reasoning that ends in one answer that can be checked exactly, and
MMLU-Pro's chains run about twice as long (median 398 against 212 completion tokens at
temperature 0.7). Each task uses 50 fixed items, the first 50 of its test split.

\textbf{Grading.} Each generation has a token budget, 512 tokens on GSM8K and 1024 on MMLU-Pro.
A generation that is still running when it reaches the budget is cut off there, and we call it
\emph{capped}. The parsers follow lm-eval-harness \citep{gao2023evalharness}, the framework
behind most published scores on these tasks, so our grading matches the scores readers know and
no parsing choice of ours can favour one configuration. The parser records one of three
outcomes. A generation is \emph{strict} when it contains the final-answer sentence the prompt
asks for (``The final answer is 26''; ``The answer is (I)''). It is \emph{flexible} when that
sentence is absent and the parser falls back to the last number or the last parenthesized
letter. It is \emph{failed} when neither exists, and it is then graded incorrect.
Appendix~\ref{app:grading} shows one generation of each kind.

\textbf{Models.} The main grid holds seven instruction-tuned models from five families:
Llama-3.1-8B \citep{grattafiori2024llama}, Mistral-7B-v0.3 \citep{jiang2023mistral}, Qwen2.5-7B
\citep{yang2024qwen25}, Qwen3-1.7B, 4B, and 8B with thinking off \citep{yang2025qwen3}, and
Gemma-3-12B \citep{gemma2025gemma3}. They are the widely used open-weight families at sizes that
fit a single consumer GPU with at most 16\,GB of memory (Appendix~\ref{app:repro}). Each
runs at four quantization levels, Q8\_0, Q6\_K, Q4\_K\_M, and Q3\_K\_M, where the digit is the
number of bits per weight, all produced by one provider with the same conversion settings
(Appendix~\ref{app:repro}), so the levels differ only in precision. Every temperature result in
the main text uses Q8\_0, the level closest to full precision. Six more models run the temperature
test at Q8\_0. Two are further Llama-3 releases, Llama-3-8B \citep{meta2024llama3} and
Llama-3.2-3B \citep{meta2024llama32}, added because Llama-3.1-8B collapsed in the grid and we
wanted to know whether the other releases of the family do. Hermes-3-8B \citep{teknium2024hermes3}
is a different post-training of the Llama-3.1-8B base, which separates the base model from its
post-training. Qwen3.5-9B \citep{qwen2026qwen35} and Gemma-4-E4B \citep{google2026gemma4} are
two releases from 2026, added to test whether current models show the same pattern, and
OLMo-3-7B \citep{ai2025olmo3} is a fully open model. Appendix~\ref{app:repro} gives the model
files, their hashes, and the llama.cpp builds.

\textbf{Fragility flag.} Hermes-3, Qwen3.5-9B, Gemma-4-E4B, and OLMo-3-7B run the temperature
test alone first. A model then receives the eight decoding configurations only if the test shows
that it loses accuracy to temperature, because the sampler comparison on models that hold their
accuracy is already covered by the main grid. We fixed the criterion before any of the four ran.
A model qualifies when, between temperatures 0.7 and 1.3 under plain sampling, its accuracy
falls by at least 15 points, its capped rate rises by at least 20 points, or its share of strict
parses falls by at least 20 points, on either task. Three qualified, Hermes-3, Qwen3.5-9B, and
OLMo-3-7B, and each ran the eight configurations at Q8\_0. The two further Llama releases run
the temperature test only, because Llama-3.1-8B's grid already stands for the family.

\textbf{Engine check.} Every result in the grid comes from llama.cpp on quantized weights, so a
collapse could come from the engine or from the conversion of the weights. To check, we rerun
the temperature test through Hugging Face Transformers \citep{wolf2020transformers} on the
official weights: Llama-3.2-3B at full 16-bit precision, Llama-3.1-8B in 8-bit weights, the
largest that fit the GPU's 16\,GB, and Qwen3-4B at full precision as a model that holds its accuracy.

\textbf{Statistical inference.} All intervals in the paper are 95\% confidence intervals
obtained by bootstrapping over items. We resample the 50 items in each task, keeping all
generations of an item together. This captures uncertainty due to the choice of items without
treating repeated generations as independent observations. Comparisons between configurations
are paired because each configuration is evaluated on the same 50 items. In
Section~\ref{sec:bound}, we ask whether any truncation sampler improves on plain temperature
sampling. To account for testing several samplers, we resample their differences from
temperature sampling jointly and compute an upper confidence bound on the largest improvement
\citep{westfall1993resampling}.

\begin{figure*}[t]
\centering
\includegraphics[width=0.92\textwidth]{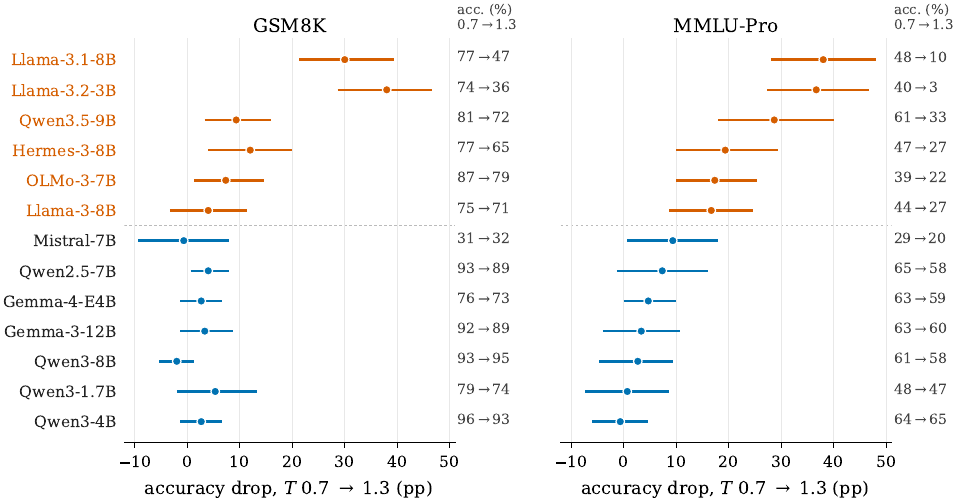}
\caption{Accuracy drop under plain temperature sampling from $T{=}0.7$ to $1.3$ at Q8\_0, with
95\% item-clustered bootstrap intervals; the accuracies at the two temperatures are printed at
the right. Positive values are losses. Models are sorted by the MMLU-Pro drop; the six in orange above the
dotted line lose at least 15 points on one task, the accuracy criterion of the fragility flag.}
\label{fig:drops}
\end{figure*}

\section{Results}

\subsection{Temperature sensitivity differs by model}
\label{sec:sensitivity}

Figure~\ref{fig:drops} gives the accuracy drop from temperature 0.7 to 1.3 under plain sampling
for all thirteen models at Q8\_0, with the accuracies at every temperature in
Appendix~\ref{app:temp}. The models divide into two groups. Six lose between 17 and 38 points
on MMLU-Pro. The other seven models lose at most 10 points, and for each of these seven the
confidence interval includes a drop of 4 points or less. We call the first group fragile and
the second robust. On GSM8K, the shorter task, only Llama-3.1-8B and Llama-3.2-3B lose heavily,
30 and 38 points, and every other model loses 12 points or less.

The collapse survives a change of engine and precision. Run through Hugging Face Transformers
on its official 16-bit weights, Llama-3.2-3B loses 40 points on GSM8K and 36 on MMLU-Pro,
against 38 and 37 through llama.cpp at Q8\_0, a difference within the confidence intervals on
both tasks. Llama-3.1-8B in 8-bit weights collapses as well, and Qwen3-4B at full precision
holds its accuracy as it does at Q8\_0 (Appendix~\ref{app:repro}). Post-training changes the
size of the collapse. Hermes-3-8B shares the Llama-3.1-8B base and differs from it only in
post-training, and it loses half as much on MMLU-Pro, 19 points against 38, with half the rise
in capped generations, 38 points against 75.

\subsection{The lost accuracy is degenerate output}
\label{sec:degeneration}

\begin{figure}[t]
\centering
\includegraphics[width=\columnwidth]{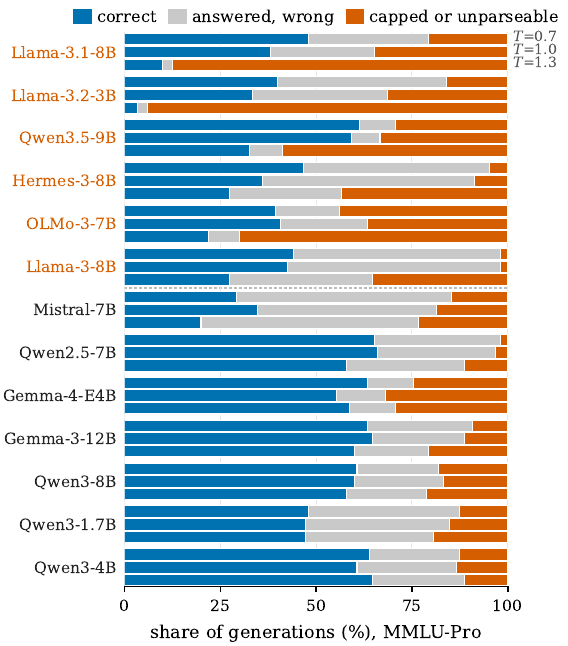}
\caption{Outcome of every MMLU-Pro generation under plain temperature sampling at Q8\_0, three
bars per model, top to bottom $T{=}0.7$, 1.0, and 1.3. Categories are exclusive, in this
precedence: correct; capped or unparseable (so this category holds incorrect generations only);
answered wrong. Model order as in Figure~\ref{fig:drops}.}
\label{fig:outcomes}
\end{figure}

An accuracy drop can come from two kinds of failure. Either the model completes an answer and
the answer is wrong, or the model never reaches an answer. The parser separates the two.
Figure~\ref{fig:outcomes} divides every MMLU-Pro generation under plain temperature sampling
into three outcomes. The first is correct. The second is capped or unparseable: the generation
reached its token budget, or it contained no answer the parser could grade. The third is
answered wrong: the generation ended before the budget with a parsed answer, and the answer is
wrong. In all six fragile models the capped-or-unparseable share rises between temperatures 0.7
and 1.3, by 26 to 78 points, while the answered-wrong share falls or stays flat. In the seven
robust models the capped-or-unparseable share moves by at most 11 points. GSM8K shows the same
pattern for Llama-3.1-8B and Llama-3.2-3B (Appendix~\ref{app:grading}), with one difference in
grading. Incoherent text usually contains a number, so on GSM8K the parser's fallback grades it
as a wrong answer, while on MMLU-Pro it fails to parse. The capped flag records the failure on
both tasks. Among the generations that do complete at 1.3, accuracy is within a point of greedy
decoding on the same items for Llama-3.1, Qwen3.5, and OLMo, and below it for the other three
fragile models (Appendix~\ref{app:grading}).

Capped generations at 0.7 and at 1.3 fail in different ways. To see what they contain, we
sampled 152 generations from four models across the three outcomes at temperatures 0.7 and 1.3
and had Claude Fable 5.1 \citep{anthropic2026fable} label each one as completed, misformatted,
cut off while coherent, degenerate, or other. The labelling model saw only the task and the raw text,
and it did not know which model or temperature had produced the text. Every audited Llama-3.1
generation capped at 1.3 was labelled degenerate. Every one capped at 0.7 was labelled cut off
while coherent. The second kind is common at 0.7 in verbose models: OLMo-3 and Qwen3.5 reach
the budget on 47 and 29\% of their MMLU-Pro generations there.

\begin{figure*}[t]
\centering
\includegraphics[width=0.95\textwidth]{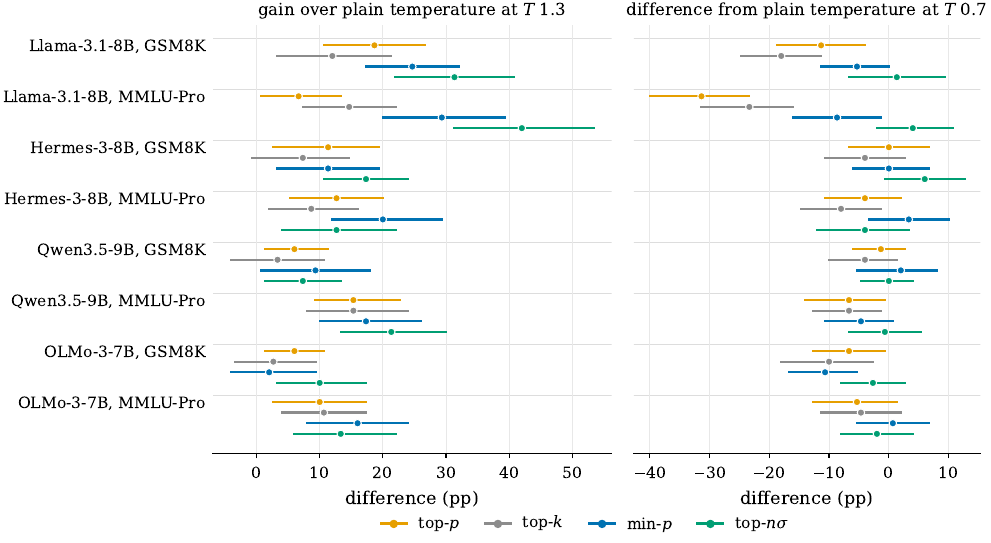}
\caption{Truncation at $T{=}1.3$ on the four fragile checkpoints with a sampler grid
(Llama-3.1-8B at Q8\_0 from the main grid; the three flagged panel models). Samplers at their standard
settings with temperature applied first. Left: paired difference from plain temperature at
1.3. Right: paired difference from plain temperature at 0.7. Pointwise 95\% item-clustered
bootstrap intervals. Every difference in the left panel is positive, and on the three flagged
models 20 of the 24 intervals exclude zero. For the flagged models both plain-temperature
baselines are rerun within the sampler-grid run and differ from the extension run's by up to 6
points.}
\label{fig:recovery}
\end{figure*}
\subsection{No sampler gains where accuracy holds}
\label{sec:bound}

On the six robust models with a sampler grid, no truncation sampler improves on plain
temperature sampling at temperatures 0.7 and 1.0. Table~\ref{tab:bound} summarizes the
comparisons. Each truncation sampler is compared with plain temperature sampling on the same
items, for every robust model in the main grid, both tasks, and both temperatures. Almost every
difference lies within 3 points of zero, and the largest gain is 3.3 points. Because the
question is whether any of these many comparisons shows a gain, the table also gives an upper
confidence bound on the largest gain across all of them, which is 9.0 points. The bound is a
limit on how much a truncation sampler could gain on these models and tasks at these
temperatures.

The first gains appear where accuracy has begun to fall. On the same six models at temperature
1.3, gains reach 9 points, and the comparisons with intervals above zero belong to Mistral-7B,
Qwen2.5-7B, and Gemma-3-12B, the three robust models with the largest drops. Llama-3.1-8B, the
fragile model in the main grid, shows a gain already at 1.0. Its MMLU-Pro accuracy, pooled over
quantization levels, has fallen 8 points by that temperature, and min-$p$ and top-$n\sigma$ each
gain about 11.
Reversing the order of temperature scaling and truncation changes accuracy by at most 4 points
on the main-grid models at 0.7 and 1.0 (Appendix~\ref{app:sampler}).

\begin{table*}[t]
\centering
\caption{Paired gains of the four truncation samplers over plain temperature sampling, by
family of comparisons. For each family: the number of comparisons, the largest gain, the 95\%
upper confidence bound on the largest gain (the 95th percentile of the jointly resampled
maximum, not computed for the robust models at 1.3, where the gains are the finding), and the number of pointwise
intervals above zero. The six robust models are Mistral-7B, Qwen2.5-7B, Qwen3-1.7B, Qwen3-4B,
Qwen3-8B, and Gemma-3-12B.}
\label{tab:bound}
\footnotesize\setlength{\tabcolsep}{4pt}\begin{tabular}{llrrrr}
\toprule
Models & $T$ & comparisons & largest gain (pp) & upper bound (pp) & intervals above 0 \\
\midrule
six robust models, all quantization levels & 0.7, 1.0 & 96 & $+3.3$ & 9.0 & 1 \\
six robust models, Q8\_0 only & 0.7, 1.0 & 96 & $+4.7$ & 16.0 & 0 \\
six robust models, all quantization levels & 1.3 & 48 & $+9.0$ & -- & 13 \\
Llama-3.1-8B, all quantization levels & 0.7, 1.0 & 16 & $+11.2$ & 16.8 & 6 \\
three flagged models, Q8\_0 & 0.7, 1.0 & 48 & $+8.0$ & 18.0 & 2 \\
three flagged models, Q8\_0 & 1.3 & 24 & $+21.3$ & 32.7 & 20 \\
\bottomrule
\end{tabular}

\end{table*}

\begin{figure*}[t]
\centering
\includegraphics[width=0.9\textwidth]{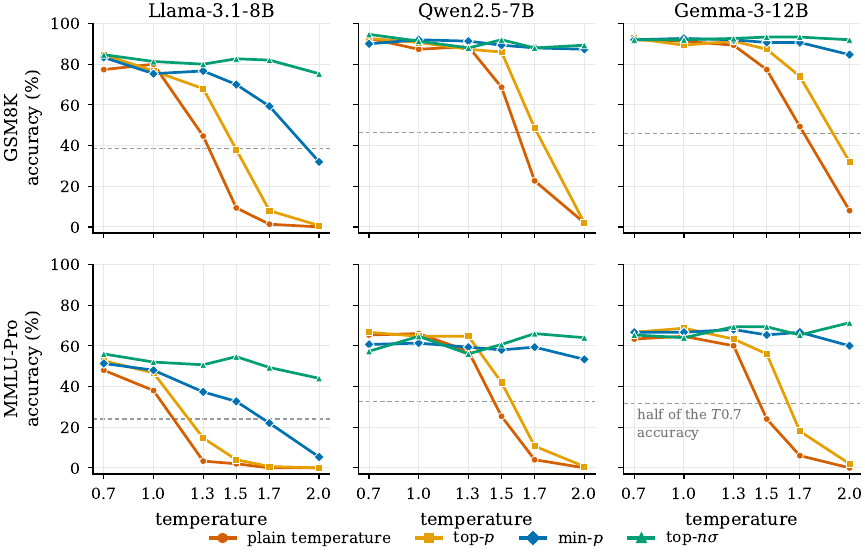}
\caption{Accuracy against temperature under four configurations at Q8\_0 (points at 0.7 and 1.0
from the main grid, 1.3 to 2.0 from the temperature sweep). The dashed line marks half of the same
model and task's plain-temperature accuracy at 0.7 in the main grid, the collapse criterion;
crossings are read from point estimates on the tested temperatures.}
\label{fig:ladder}
\end{figure*}
\subsection{Truncation recovers accuracy on the collapse}
\label{sec:recovery}

On the fragile models, every truncation sampler improves accuracy at temperature 1.3
(Figure~\ref{fig:recovery}). The gains are large because the collapse is large. On every model
and task the best sampler brings accuracy back to within six points of its level at 0.7, so the
accuracy lost between 0.7 and 1.3 under plain sampling was almost entirely recoverable by
removing the tail of the distribution. Which sampler recovers the most varies from model to
model, and because each sampler was compared with plain temperature sampling only, we do not
rank them.

\textbf{Where each model collapses.} The robust models collapse too, at higher temperatures.
Figure~\ref{fig:ladder} extends the temperature to 2.0 on Llama-3.1-8B, Qwen2.5-7B, and
Gemma-3-12B under plain sampling and three truncation samplers. A model has collapsed once its
accuracy falls below half of what it was at 0.7, and the collapse temperature is the first
tested temperature at which that happens. Under plain sampling the three models collapse in the
same order as in Figure~\ref{fig:drops}, Llama-3.1 first and Gemma-3 last, and each collapses
earlier on MMLU-Pro, the longer task, than on GSM8K. Truncation moves the collapse to a higher
temperature or removes it from the tested range. Top-$p$ delays it in half of the model and task
combinations, min-$p$ delays it further, and top-$n\sigma$ does not collapse on any model or
task through 2.0. Min-$p$ and top-$n\sigma$ were validated at temperatures of 1.5 to 3
\citep{nguyen2024minp,tang2024topn}, the range in which these collapses occur, and that is where
their gains are largest here as well.

\section{Discussion and conclusion}
\label{sec:discussion}

We asked under what model and temperature conditions truncation samplers improve accuracy over
plain temperature sampling. The answer depends on whether temperature has already reduced a
model's accuracy. Truncation helps when accuracy has declined, and the larger the decline, the
larger the gain. When accuracy remains stable, truncation offers no improvement. At
temperatures used in practice, twelve of the thirteen models maintained their accuracy. The
sole exception, Llama-3.1-8B at temperature 1.0, was also the only model that benefited from
truncation.

This pattern reconciles findings in the sampler literature with practical experience. Min-$p$
and top-$n\sigma$ were evaluated at temperatures of 1.5 to 3
\citep{nguyen2024minp,tang2024topn}, a range in which accuracy collapsed for every model in our
sweep and truncation helped. At temperature 1.0, by contrast, the Min-$k$ paper reports that
top-$k$, top-$p$, and min-$p$ perform within one percentage point of one another
\citep{ding2026mink}. That finding is consistent with ours, while the paper's broader claim
that sampler choice matters across the temperature range rests on results above temperature 1.

For practitioners working on tasks with checkable answers, model and temperature should be
chosen together, followed by the sampler. If the model maintains its accuracy at the intended
temperature, a standard truncation sampler makes little difference. If accuracy declines,
either lowering the temperature or applying truncation recovers most of the loss.

This sensitivity to temperature can be detected without gold-standard answers. Neither the
capped rate nor the strict-parse rate requires labels, and changes in these rates between
temperatures 0.7 and 1.3 distinguished fragile models from robust ones in every run of this
study, including the four models tested after the criterion was fixed. Our evaluation measured
accuracy only, so it does not establish what gains in output diversity higher temperatures
might offer.

Evaluations of new samplers should report accuracy gains alongside the accuracy lost under
plain temperature sampling at the same temperature. A sampler that helps one model after its
accuracy has collapsed may offer no benefit to a model that remains robust at that temperature.

What determines a model's collapse temperature remains an open question. The Hermes-3 results
show that post-training changes the magnitude of the collapse for one base model. However, our
study includes neither a base-versus-instruct comparison nor a series of checkpoints, so it
cannot identify which aspects of post-training drive this change.

Truncation samplers were designed to preserve coherence at high temperatures. On tasks with
checkable answers, their benefit follows the same pattern: they recover accuracy lost to
temperature but offer no improvement when accuracy remains stable. At temperatures used in
practice, model choice determines accuracy.

\section*{Limitations}

Each task uses 50 fixed items with three repetitions. The small item count is what makes a
factorial run of 185,600 generations feasible, and every estimate carries an interval that
reflects it. The 50 MMLU-Pro items are the first 50 of the test split, which is sorted by
category, so all of them are business questions. A second set of 50 items drawn from all 14
categories reproduces Llama-3.1's much larger drop on three models (Appendix~\ref{app:temp}),
but as another sample of 50 it does not estimate the effect on the whole benchmark. Mistral-7B
scores about 30\% on both tasks at 0.7, which leaves little accuracy to lose and limits what its
robust classification means.

Temperature 1.3 was chosen as a single point above the defaults to provide an upper bound on
the range used in practice. The study therefore does not establish where between 1.0 and 1.3
each model's accuracy begins to collapse. Token budgets were fixed, and verbose models reached
them even at temperature 0.7 for reasons unrelated to temperature. The audit distinguishing
degenerate text from reasoning cut short by the token budget covered 152 generations from four
models in the main grid, and 34 labels were marked as borderline. Generations from the three
flagged models were not audited, so their capped rates serve as a proxy for degeneration.

The sampler results rest on fewer models than the temperature results. Ten models ran the
sampler comparison, and three of them were included because they had already collapsed, so the
recovery in Section~\ref{sec:recovery} describes models that were chosen for being fragile.
Within each model and task, the best sampler is the one that scored highest, identified after
seeing the results, so a sampler chosen in advance would recover less. Each sampler ran at one
setting, the one its source paper recommends, and the upper bound of Section~\ref{sec:bound}
applies to those settings only.

The scope is open-weight instruction-tuned models of 1.7 to 12 billion parameters, in English,
zero-shot, on tasks with one checkable answer, with thinking mode off. A spot check of Qwen3-8B
with thinking on shows a 10-point drop on MMLU-Pro at 25 items (Appendix~\ref{app:temp}), so
the results do not extend to thinking mode. Open-ended generation, the original use of
truncation samplers, lies outside the design, as does the output diversity that a higher
temperature buys when several samples are drawn and the best is selected. The study identifies
no cause for the differences between models.

\bibliography{references}

\appendix

\section{Reproducibility and engine check}
\label{app:repro}

\textbf{Builds and hardware.} Every run in the main grid, the temperature test, and the sweep
went through llama.cpp \citep{gerganov2023llamacpp} on a consumer GPU, an RTX 2080~Ti or an
RTX 5070~Ti with 16\,GB, with a CUDA build for each architecture. The main grid uses commit
\texttt{5aba5364} (server build 9456). Qwen3.5-9B and Gemma-4-E4B require a later release, so
the six additional models, their sampler grids, and a bridge run use release v0.4.0 (commit
\texttt{5266f24d}). The bridge reruns the temperature test of Llama-3.2-3B and Qwen3-4B on
v0.4.0. Both reproduce the drops measured on the main-grid build, with differences whose
intervals include zero (Table~\ref{tab:engine-full}), so results from the two builds are
compared directly in the paper.

\textbf{Prompts.} The client renders each model's official chat template and sends the
resulting text, with the server's own template parser and beginning-of-sequence insertion
turned off, so the client owns every special token. Each record stores the SHA-256 hash of the
rendered prompt. Release v0.4.0 inserts its own beginning-of-sequence token for Gemma~4, so the
Gemma-4 prompt is rendered without one. The Llama template stamps the run date into the system
header, which is the only difference between the prompt strings of runs made on different
days; the prompt token counts agree on every item.

\textbf{Zero-token records.} On both builds the server occasionally returns a degenerate
generation of budget length with a token count of zero. The main-grid build returns the text,
which the client recovers and which the audit of Appendix~\ref{app:grading} labelled;
v0.4.0 returns an HTTP error without text. The runner records both as capped generations with
zero tokens. They number 189 of the 185,600 records in the main grid, 114 in the run of the two
further Llama releases, 158 in the temperature test of the four 2026 models, 196 in their
sampler grids, 64 in the engine check, 73 in the bridge, and 2,549 in the sweep. At $T{=}1.3$
under plain temperature sampling they are 13\% of Llama-3.1's capped generations on MMLU-Pro, 44\%
of Llama-3.2-3B's, and 37 of Llama-3-8B's 44.

\textbf{Model files.} Every quantized model is a Bartowski imatrix GGUF from the same provider,
so the four quantization levels of a model differ only in precision. The SHA-256 hash of every
model file is listed in the repository (\texttt{DOWNLOADS.md}).

\textbf{Decoding and seeding.} Each request carries exactly one truncation parameter, and the
order of temperature scaling and truncation is pinned through the server's \texttt{samplers}
array and recorded in every record. On both builds the server instantiates only the samplers
named in that array, so its defaults for the samplers not named (top-$k$ 40, top-$p$ 0.95,
min-$p$ 0.05) have no effect, and no repetition penalty or other logit processor is sent. The
seed of every generation is derived from the global seed 0 and the model, quantization,
sampler, temperature, task, item, and repetition. The server serves up to eight requests at
once, so a run is seed-logged but not bit-exact. The tasks are GSM8K (\texttt{openai/gsm8k},
test split, first 50 items) and MMLU-Pro (\texttt{TIGER-Lab/MMLU-Pro}, test split, first 50
items). Each generation is one JSONL record with its identity, request parameters, raw output,
parse outcome, and token counts, and runs resume by record id. The pipeline, configurations,
analysis scripts, and every record are at
\url{https://github.com/larosafrancesco289/decoding-robustness}.

\textbf{The two further Llama releases.} Llama-3-8B and Llama-3.2-3B ran the temperature test
before the fragility criterion of Section~\ref{sec:design} was written. Both meet it. They did
not run the eight decoding configurations, and Llama-3.1-8B's grid stands for the family in
Sections~\ref{sec:bound} and \ref{sec:recovery}.

\textbf{Engine check.} The Transformers \citep{wolf2020transformers} configuration loads
\texttt{meta-llama/Llama-3.2-3B-Instruct} (revision \texttt{0cb88a4f}) in BF16 with
transformers 5.16.1 and torch 2.14. It samples with \texttt{do\_sample} at temperature $T$,
with top-$k$ and top-$p$ disabled and no other processor, in batches of eight with left
padding; greedy decoding is \texttt{do\_sample=False}. Generation stops at the model's
end-of-sequence tokens or at the task's budget. Llama-3.1-8B loads the
\texttt{unsloth/Llama-3.1-8B-Instruct} mirror of the official weights (revision
\texttt{4699cc75}) in 8-bit through bitsandbytes, in batches of four. Qwen3-4B loads
\texttt{Qwen/Qwen3-4B} (revision \texttt{1cfa9a72}) in BF16 with thinking disabled. All three
reproduce every grid prompt hash. Llama-3.2-3B also ran through llama.cpp on its unquantized
16-bit GGUF, which separates the engine from the quantization. Table~\ref{tab:engine-full}
gives, for every configuration, the accuracy at 0.7 and 1.3, the drop with its interval, the
capped rate at 1.3, and the paired difference of the drop from the main-grid Q8\_0 cell. The
intervals on those differences are about 20 points wide, so the check shows that the collapse
recurs, and Llama-3.1's GSM8K drop is 13 points larger in 8-bit weights than at Q8\_0.

\begin{table*}[t]\centering
\caption{Engine, precision, and build checks: plain-temperature accuracy at $T{=}0.7$ and
$1.3$, the drop with its 95\% item-clustered interval, the capped rate at 1.3 (\%), and the
difference of the drop from the reference Q8\_0 llama.cpp cell on the same items (positive: a
larger drop than the reference). The BF16 GGUF row of Qwen3-4B belongs to the main grid.}
\label{tab:engine-full}
\footnotesize\setlength{\tabcolsep}{3pt}\resizebox{\textwidth}{!}{\begin{tabular}{llcclclcclcl}
\toprule
 & & \multicolumn{5}{c}{GSM8K} & \multicolumn{5}{c}{MMLU-Pro} \\
Model & Precision, engine & 0.7 & 1.3 & drop [CI] & cap & vs.\ ref.\ [CI] & 0.7 & 1.3 & drop [CI] & cap & vs.\ ref.\ [CI] \\
\midrule
Llama-3.2-3B & BF16, transformers & 74.7 & 34.7 & $+40.0$ [29.3, 49.3] & 61 & $+2.0$ [-8.7, 12.0] & 40.7 & 4.7 & $+36.0$ [24.7, 48.0] & 92 & $-0.7$ [-9.3, 8.7] \\
 & F16 GGUF, llama.cpp & 74.0 & 42.7 & $+31.3$ [22.7, 40.7] & 53 & $-6.7$ [-17.3, 4.0] & 38.0 & 5.3 & $+32.7$ [23.3, 43.3] & 91 & $-4.0$ [-10.7, 2.7] \\
 & Q8 GGUF, llama.cpp v0.4.0 & 77.3 & 34.7 & $+42.7$ [34.0, 50.7] & 63 & $+4.7$ [-3.3, 12.7] & 39.3 & 4.7 & $+34.7$ [24.7, 45.3] & 93 & $-2.0$ [-9.3, 5.3] \\
 & Q8 GGUF, llama.cpp (reference) & 74.0 & 36.0 & $+38.0$ [28.7, 46.7] & 55 & -- & 40.0 & 3.3 & $+36.7$ [27.3, 46.7] & 92 & -- \\
\midrule
Llama-3.1-8B & INT8, transformers & 82.7 & 39.3 & $+43.3$ [33.3, 52.0] & 54 & $+13.3$ [2.7, 23.3] & 45.3 & 2.7 & $+42.7$ [32.7, 53.3] & 95 & $+4.7$ [-3.3, 12.7] \\
 & Q8 GGUF, llama.cpp (reference) & 77.3 & 47.3 & $+30.0$ [21.3, 39.3] & 51 & -- & 48.0 & 10.0 & $+38.0$ [28.0, 48.0] & 86 & -- \\
\midrule
Qwen3-4B & BF16, transformers & 94.7 & 93.3 & $+1.3$ [-1.3, 4.0] & 0 & $-1.3$ [-5.3, 2.0] & 60.7 & 61.3 & $-0.7$ [-6.0, 5.3] & 13 & $+0.0$ [-6.7, 6.7] \\
 & BF16 GGUF, llama.cpp & 95.3 & 96.7 & $-1.3$ [-4.0, 1.3] & 0 & $-4.0$ [-9.3, 1.3] & 58.0 & 60.0 & $-2.0$ [-7.3, 3.3] & 15 & $-1.3$ [-8.7, 6.0] \\
 & Q8 GGUF, llama.cpp v0.4.0 & 96.7 & 94.0 & $+2.7$ [-0.7, 7.3] & 0 & $+0.0$ [-4.0, 4.7] & 65.3 & 64.0 & $+1.3$ [-4.7, 7.3] & 11 & $+2.0$ [-6.0, 10.0] \\
 & Q8 GGUF, llama.cpp (reference) & 96.0 & 93.3 & $+2.7$ [-1.3, 6.7] & 1 & -- & 64.0 & 64.7 & $-0.7$ [-6.0, 4.7] & 12 & -- \\
\bottomrule
\end{tabular}
}
\end{table*}

\section{Grading and degeneration}
\label{app:grading}

\begin{figure}[t]
\centering
\includegraphics[width=0.93\columnwidth]{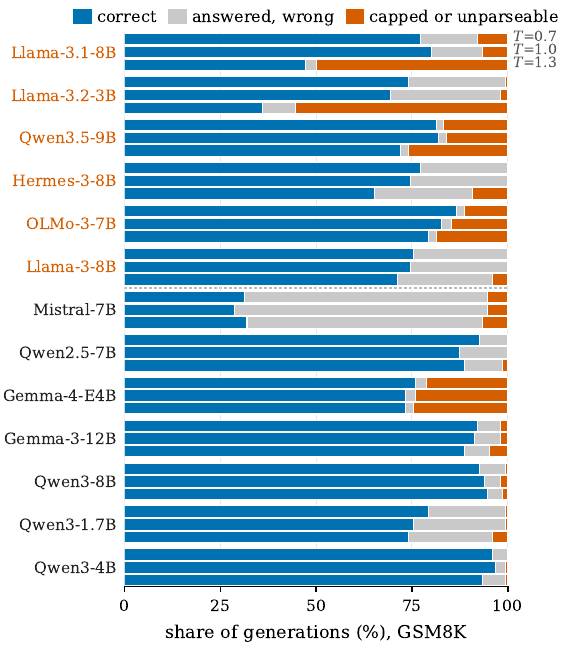}
\caption{Outcome of every GSM8K generation under plain temperature sampling at Q8\_0, as in
Figure~\ref{fig:outcomes}. On GSM8K a degenerate generation usually contains a number, so part
of the degeneration is graded through the flexible path; the capped flag records it.}
\label{fig:outcomes-gsm8k}
\end{figure}

\textbf{Parse outcomes.} Figure~\ref{fig:parsepaths} shows one generation for each of the
three parse outcomes of Section~\ref{sec:design}, and Figure~\ref{fig:degenexamples} two
degenerate generations. The strict outcome requires the final-answer sentence with the answer
directly after it. The flexible outcome, the parser's fallback to the last number or the last
parenthesized letter, catches two different things: a well-formed answer written outside that
sentence, and an incoherent generation that still contains a number, which on GSM8K is graded
as a wrong answer. Mistral-7B omits the sentence on 26 to 32\% of its GSM8K generations at
every temperature, and the Qwen3 models wrap the answer in markdown emphasis, which the strict
pattern does not match; both grade correctly through the fallback. To check that markdown does
not distort the results, we re-graded every main-grid record with a strict pattern that
tolerates it: 598 of the 185,600 grades change, 556 of them on Qwen3-8B and Qwen3-4B MMLU-Pro,
and no plain-temperature drop moves by more than 1.0 point. The original parser, fixed before
any run, stays primary. Table~\ref{tab:decomp} gives the share of each outcome and the capped
rate for every model and temperature, and Figure~\ref{fig:outcomes-gsm8k} the GSM8K outcomes
in the form of Figure~\ref{fig:outcomes}.

\begin{figure}[t]
\centering
\begin{minipage}{0.96\columnwidth}
\hrule\vspace{4pt}
{\small\ttfamily\raggedright
[strict] \ldots{} x = 19.50 / 0.75 \quad x = 26 \quad So the original price of the book
was \$26. \quad The final answer is \$26.
\par}
\vspace{4pt}\hrule\vspace{4pt}
{\small\ttfamily\raggedright
[flexible] \ldots{} 80 \textbackslash{}text\{ miles\} + 150 \textbackslash{}text\{ miles\}
= 230 \textbackslash{}text\{ miles\} \quad \#\#\# Final Answer: \quad The final answer is
**230**.
\par}
\vspace{4pt}\hrule\vspace{4pt}
{\small\ttfamily\raggedright
[failed] \ldots{} \textbackslash{}frac\{3,462.20\}\{35\} = 98.92 \quad \#\#\# Final
Answer: \quad The answer is **J**.
\par}
\vspace{4pt}\hrule
\end{minipage}
\caption{One generation per parse outcome (endings only). Strict: Llama-3.1-8B, GSM8K,
$T{=}1.3$, graded 26, correct. Flexible: Qwen3-8B, GSM8K, $T{=}0.7$; the emphasis markers defeat
the strict pattern and the last-number fallback grades 230, correct. Failed: Qwen3-8B, MMLU-Pro,
$T{=}0.7$; the parser expects a parenthesized letter such as (J), none appears, graded
incorrect.}
\label{fig:parsepaths}
\end{figure}

\begin{figure}[t]
\centering
\begin{minipage}{0.96\columnwidth}
\hrule\vspace{5pt}
{\small\ttfamily\raggedright
[opens] To solve this problem, we'll use the gross profit method of inventory
evaluation. Step 1: Calculate the gross profit earned during the year\ldots\\{}
[ends] H sieve Con Palette sediment conduit France called coaching OMAX may PHYS today
Enhanced vertices XOR/t and registration ordinances efforts yesterday Michael
\par}
\vspace{5pt}\hrule\vspace{5pt}
{\small\ttfamily\raggedright
[opens] To determine the correct type of research methods, we need to understand the
characteristics of each option: A. Non-probability: This method involves
selecting\ldots\\{}
[ends] entageeba lever modeling ign measures inherits cones quantum auditing\-ippines
image later campaigns Taylorart parte diesel El N resembled Dental professionals
\par}
\vspace{5pt}\hrule
\end{minipage}
\caption{Two degenerate generations (Llama-3.1-8B, Q8\_0, plain temperature, $T{=}1.3$,
MMLU-Pro; both reached the 1024-token budget). Each opens coherently, loses coherence
mid-generation, and continues as token noise until the budget.}
\label{fig:degenexamples}
\end{figure}

\textbf{Blinded audit.} To check that the parse outcomes mean what Section~\ref{sec:degeneration}
takes them to mean, we drew 152 plain-temperature generations at Q8\_0 from Llama-3.1-8B,
Mistral-7B, Qwen2.5-7B, and Qwen3-8B at temperatures 0.7 and 1.3 on both tasks, stratified by
parse outcome with the uncertain outcomes oversampled. Claude Fable 5.1
\citep{anthropic2026fable} labelled each generation as completed, cut off while coherent,
degenerate, misformatted, or other, seeing only the task and the raw text. Of the 51 capped
generations, all 8 of Llama-3.1's at 1.3 are degenerate and all 8 at 0.7 are coherent reasoning
cut short; the robust models' 21 capped generations at 1.3 split 8 degenerate, 9 cut off, 3
completed, and 1 misformatted, and their 14 at 0.7 split 13 cut off and 1 completed. Of the 26
parse failures, 15 contain a clear answer in an unexpected form, 6 are degenerate, 4 are
refusals, and 1 is cut off. Of the 75 strict and flexible parses, 72 are completed answers. The
annotator marked 34 of the 152 labels as borderline. The sample, the labels, and the key are in
the repository (\texttt{results/outcome\_audit}).

\textbf{Surviving answers at 1.3.} Among the generations that do complete with the
final-answer sentence at 1.3, accuracy is subject to selection, because the surviving
generations can come from an easier mix of items. Table~\ref{tab:survivor} therefore gives, for
every model, the share of plain-temperature generations with a strict parse at 0.7 and 1.3, the
accuracy among them, and greedy accuracy on the same surviving items, each survivor counted
once. On MMLU-Pro at 1.3, Llama-3.1's completed answers score 78.9\% against 78.9\% for greedy
decoding on the same items, Qwen3.5's 83.1 against 84.7, and OLMo-3's 73.3 against 73.3. For
Llama-3-8B, Llama-3.2-3B (nine survivors), and Hermes-3 the completed answers score below greedy
decoding by 4, 11, and 16 points, and so do the robust Mistral-7B's by 21 points, so a gap
among the survivors at 1.3 is not specific to fragility. The comparison controls for the mix of
items and does not remove the selection.

\begin{table*}[t]\centering
\caption{Parse outcomes under plain temperature sampling at Q8\_0 (\% of generations,
rounded): strict, flexible, failed, and capped (capped overlaps the other three).}
\label{tab:decomp}
\scriptsize\renewcommand{\arraystretch}{0.95}\setlength{\tabcolsep}{5pt}\begin{tabular}{llrrrrrrrr}
\toprule
 & & \multicolumn{4}{c}{GSM8K} & \multicolumn{4}{c}{MMLU-Pro} \\
Model & $T$ & strict & flex. & failed & capped & strict & flex. & failed & capped \\
\midrule
Llama-3.1-8B & 0.7 & 89 & 11 & 0 & 9 & 79 & 1 & 19 & 11 \\
 & 1.0 & 87 & 13 & 0 & 7 & 63 & 5 & 32 & 22 \\
 & 1.3 & 45 & 55 & 0 & 51 & 13 & 6 & 81 & 86 \\
Llama-3.2-3B & 0.7 & 99 & 1 & 0 & 1 & 83 & 3 & 15 & 10 \\
 & 1.0 & 89 & 11 & 0 & 2 & 65 & 5 & 30 & 22 \\
 & 1.3 & 41 & 59 & 0 & 55 & 6 & 9 & 85 & 92 \\
Qwen3.5-9B & 0.7 & 80 & 20 & 0 & 21 & 71 & 3 & 26 & 29 \\
 & 1.0 & 79 & 21 & 0 & 21 & 67 & 5 & 27 & 33 \\
 & 1.3 & 71 & 29 & 0 & 27 & 39 & 9 & 51 & 43 \\
Hermes-3-8B & 0.7 & 68 & 32 & 0 & 0 & 93 & 2 & 5 & 0 \\
 & 1.0 & 68 & 32 & 0 & 0 & 91 & 1 & 9 & 0 \\
 & 1.3 & 59 & 39 & 2 & 9 & 57 & 2 & 41 & 38 \\
OLMo-3-7B & 0.7 & 84 & 16 & 0 & 14 & 56 & 11 & 33 & 47 \\
 & 1.0 & 84 & 16 & 0 & 16 & 62 & 10 & 28 & 38 \\
 & 1.3 & 78 & 17 & 5 & 22 & 30 & 0 & 70 & 68 \\
Llama-3-8B & 0.7 & 98 & 2 & 0 & 0 & 97 & 1 & 2 & 0 \\
 & 1.0 & 98 & 2 & 0 & 0 & 97 & 1 & 2 & 0 \\
 & 1.3 & 94 & 6 & 0 & 5 & 65 & 4 & 31 & 29 \\
Mistral-7B & 0.7 & 72 & 28 & 0 & 5 & 82 & 4 & 14 & 1 \\
 & 1.0 & 73 & 27 & 0 & 6 & 81 & 2 & 17 & 1 \\
 & 1.3 & 71 & 29 & 0 & 7 & 74 & 6 & 20 & 11 \\
Qwen2.5-7B & 0.7 & 97 & 3 & 0 & 0 & 98 & 1 & 1 & 2 \\
 & 1.0 & 93 & 7 & 0 & 0 & 96 & 1 & 3 & 3 \\
 & 1.3 & 93 & 7 & 0 & 1 & 88 & 3 & 9 & 9 \\
Gemma-4-E4B & 0.7 & 45 & 55 & 0 & 25 & 74 & 8 & 18 & 26 \\
 & 1.0 & 54 & 46 & 0 & 28 & 67 & 8 & 25 & 34 \\
 & 1.3 & 49 & 51 & 0 & 29 & 68 & 10 & 22 & 32 \\
Gemma-3-12B & 0.7 & 97 & 3 & 0 & 3 & 91 & 1 & 9 & 9 \\
 & 1.0 & 98 & 2 & 0 & 2 & 89 & 3 & 9 & 11 \\
 & 1.3 & 95 & 5 & 0 & 5 & 79 & 3 & 17 & 20 \\
Qwen3-8B & 0.7 & 45 & 55 & 0 & 1 & 82 & 1 & 17 & 12 \\
 & 1.0 & 45 & 55 & 0 & 2 & 83 & 5 & 13 & 13 \\
 & 1.3 & 47 & 53 & 0 & 2 & 78 & 5 & 17 & 18 \\
Qwen3-1.7B & 0.7 & 15 & 85 & 0 & 1 & 87 & 2 & 11 & 13 \\
 & 1.0 & 22 & 78 & 0 & 1 & 85 & 1 & 14 & 14 \\
 & 1.3 & 21 & 79 & 0 & 4 & 83 & 2 & 15 & 19 \\
Qwen3-4B & 0.7 & 96 & 4 & 0 & 1 & 89 & 0 & 11 & 10 \\
 & 1.0 & 93 & 7 & 0 & 1 & 87 & 1 & 12 & 12 \\
 & 1.3 & 96 & 4 & 0 & 1 & 88 & 1 & 11 & 12 \\
\bottomrule
\end{tabular}

\end{table*}

\begin{table*}[t]\centering
\caption{Accuracy among strict-parsed generations under plain temperature sampling at Q8\_0,
with the share of generations that parse strictly, the number $n$ of such generations at
$T{=}1.3$ (of 150), and greedy accuracy on their items with each surviving generation counted
once. Point estimates without intervals.}
\label{tab:survivor}
\scriptsize\renewcommand{\arraystretch}{0.95}\setlength{\tabcolsep}{4pt}\begin{tabular}{llrrrrrrr}
\toprule
 & & \multicolumn{3}{c}{$T{=}0.7$} & \multicolumn{4}{c}{$T{=}1.3$} \\
Model & Task & strict \% & acc$\mid$strict & greedy, same items & strict \% & $n$ & acc$\mid$strict & greedy, same items \\
\midrule
Llama-3.1-8B & GSM8K & 89 & 85.1 & 81.3 & 45 & 67 & 98.5 & 95.5 \\
 & MMLU-Pro & 79 & 60.5 & 56.3 & 13 & 19 & 78.9 & 78.9 \\
Llama-3.2-3B & GSM8K & 99 & 74.3 & 74.3 & 41 & 62 & 85.5 & 90.3 \\
 & MMLU-Pro & 83 & 47.6 & 48.4 & 6 & 9 & 55.6 & 66.7 \\
Qwen3.5-9B & GSM8K & 80 & 97.5 & 92.5 & 71 & 106 & 99.1 & 93.4 \\
 & MMLU-Pro & 71 & 86.8 & 81.1 & 39 & 59 & 83.1 & 84.7 \\
Hermes-3-8B & GSM8K & 68 & 86.3 & 83.3 & 59 & 89 & 83.1 & 82.0 \\
 & MMLU-Pro & 93 & 50.0 & 54.3 & 57 & 86 & 47.7 & 64.0 \\
OLMo-3-7B & GSM8K & 84 & 98.4 & 97.6 & 78 & 117 & 97.4 & 95.7 \\
 & MMLU-Pro & 56 & 69.0 & 61.9 & 30 & 45 & 73.3 & 73.3 \\
Llama-3-8B & GSM8K & 98 & 76.9 & 76.9 & 94 & 141 & 75.2 & 77.3 \\
 & MMLU-Pro & 97 & 45.2 & 39.7 & 65 & 98 & 41.8 & 45.9 \\
Mistral-7B & GSM8K & 72 & 33.3 & 26.9 & 71 & 106 & 32.1 & 26.4 \\
 & MMLU-Pro & 82 & 35.0 & 39.8 & 74 & 111 & 25.2 & 45.9 \\
Qwen2.5-7B & GSM8K & 97 & 94.5 & 93.8 & 93 & 139 & 91.4 & 93.5 \\
 & MMLU-Pro & 98 & 66.7 & 63.3 & 88 & 132 & 65.2 & 68.2 \\
Gemma-4-E4B & GSM8K & 45 & 98.5 & 94.0 & 49 & 74 & 93.2 & 90.5 \\
 & MMLU-Pro & 74 & 83.8 & 81.1 & 68 & 102 & 82.4 & 83.3 \\
Gemma-3-12B & GSM8K & 97 & 93.8 & 90.4 & 95 & 143 & 93.0 & 93.0 \\
 & MMLU-Pro & 91 & 69.9 & 62.5 & 79 & 119 & 75.6 & 67.2 \\
Qwen3-8B & GSM8K & 45 & 97.0 & 98.5 & 47 & 70 & 100.0 & 100.0 \\
 & MMLU-Pro & 82 & 74.0 & 70.7 & 78 & 117 & 73.5 & 74.4 \\
Qwen3-1.7B & GSM8K & 15 & 91.3 & 91.3 & 21 & 31 & 71.0 & 80.6 \\
 & MMLU-Pro & 87 & 55.0 & 55.0 & 83 & 124 & 57.3 & 54.8 \\
Qwen3-4B & GSM8K & 96 & 95.8 & 97.9 & 96 & 144 & 93.8 & 97.9 \\
 & MMLU-Pro & 89 & 72.2 & 72.2 & 88 & 132 & 73.5 & 73.5 \\
\bottomrule
\end{tabular}

\end{table*}

\section{Temperature results in full}
\label{app:temp}

Table~\ref{tab:temp-full} gives plain-temperature accuracy at all three temperatures for the
thirteen models, the drop from 0.7 to 1.3 with its interval, and the change in the capped and
strict-parse rates.

\textbf{Quantization.} Table~\ref{tab:quant-drops} recomputes the drop at each quantization
level of the main grid. Llama-3.1-8B collapses at every level, and no robust model loses more
than 12 points at any level, so quantization does not change which models lose to temperature,
in line with \citet{prasad2026quantization}. Quantization itself costs little accuracy. From
Q8\_0 to Q3\_K\_M, greedy accuracy pooled over both tasks changes by
at most 5 points on the models of 7B parameters or more, and falls by 7 and 9 points on Qwen3-4B
and Qwen3-1.7B.

\textbf{Item selection.} The fixed MMLU-Pro items are the first 50 of the test split, which is
sorted by category, so all of them are business questions. Table~\ref{tab:strat} repeats the
temperature test on a second subset of 50 items with 3 to 4 from each of the 14 categories, on
Llama-3.1-8B, Qwen2.5-7B, and Gemma-3-12B. Llama-3.1's drop stays far above the other two, and
Gemma-3's stratified drop, $+7.3$ [2.0, 12.7], has an interval that excludes zero.

\textbf{Thinking mode.} A spot check of Qwen3-8B with thinking on (Q8\_0, 25 items, two
repetitions, 450 generations) reproduces the GSM8K result (94\% at 0.7 against 96\% at 1.3) and
not the MMLU-Pro one (56 against 46\%, a 10-point drop with interval [2, 20]). With thinking on,
the longer traces make Qwen3-8B less robust on the longer task.

\begin{table*}[t]\centering
\caption{Plain-temperature accuracy at Q8\_0 for all thirteen models, the drop from $T{=}0.7$ to
$1.3$ with its 95\% item-clustered interval (bold: lower limit at or above 15, a stricter
criterion than the fragility flag), and the change in capped and strict-parse rates. Models
sorted as in Figure~\ref{fig:drops}.}
\label{tab:temp-full}
\footnotesize\setlength{\tabcolsep}{4pt}\begin{tabular}{llcccrrr}
\toprule
 & & \multicolumn{3}{c}{accuracy (\%) at $T$} & & \multicolumn{2}{c}{change 0.7$\to$1.3 (pp)} \\
Model & Task & 0.7 & 1.0 & 1.3 & drop [95\% CI] & capped & strict \\
\midrule
Llama-3.1-8B & GSM8K & 77.3 & 80.0 & 47.3 & $\mathbf{+30.0}$ [21.3, 39.3] & $+43$ & $-45$ \\
 & MMLU-Pro & 48.0 & 38.0 & 10.0 & $\mathbf{+38.0}$ [28.0, 48.0] & $+75$ & $-67$ \\
Llama-3.2-3B & GSM8K & 74.0 & 69.3 & 36.0 & $\mathbf{+38.0}$ [28.7, 46.7] & $+55$ & $-57$ \\
 & MMLU-Pro & 40.0 & 33.3 & 3.3 & $\mathbf{+36.7}$ [27.3, 46.7] & $+82$ & $-77$ \\
Qwen3.5-9B & GSM8K & 81.3 & 82.0 & 72.0 & $+9.3$ [3.3, 16.0] & $+6$ & $-9$ \\
 & MMLU-Pro & 61.3 & 59.3 & 32.7 & $\mathbf{+28.7}$ [18.0, 40.0] & $+13$ & $-31$ \\
Hermes-3-8B & GSM8K & 77.3 & 74.7 & 65.3 & $+12.0$ [4.0, 20.0] & $+9$ & $-9$ \\
 & MMLU-Pro & 46.7 & 36.0 & 27.3 & $+19.3$ [10.0, 29.3] & $+38$ & $-36$ \\
OLMo-3-7B & GSM8K & 86.7 & 82.7 & 79.3 & $+7.3$ [1.3, 14.7] & $+8$ & $-6$ \\
 & MMLU-Pro & 39.3 & 40.7 & 22.0 & $+17.3$ [10.0, 25.3] & $+21$ & $-26$ \\
Llama-3-8B & GSM8K & 75.3 & 74.7 & 71.3 & $+4.0$ [-3.3, 11.3] & $+5$ & $-4$ \\
 & MMLU-Pro & 44.0 & 42.7 & 27.3 & $+16.7$ [8.7, 24.7] & $+29$ & $-32$ \\
\midrule
Mistral-7B & GSM8K & 31.3 & 28.7 & 32.0 & $-0.7$ [-9.3, 8.0] & $+1$ & $-1$ \\
 & MMLU-Pro & 29.3 & 34.7 & 20.0 & $+9.3$ [0.7, 18.0] & $+9$ & $-8$ \\
Qwen2.5-7B & GSM8K & 92.7 & 87.3 & 88.7 & $+4.0$ [0.7, 8.0] & $+1$ & $-4$ \\
 & MMLU-Pro & 65.3 & 66.0 & 58.0 & $+7.3$ [-1.3, 16.0] & $+7$ & $-10$ \\
Gemma-4-E4B & GSM8K & 76.0 & 73.3 & 73.3 & $+2.7$ [-1.3, 6.7] & $+3$ & $+5$ \\
 & MMLU-Pro & 63.3 & 55.3 & 58.7 & $+4.7$ [0.0, 10.0] & $+6$ & $-6$ \\
Gemma-3-12B & GSM8K & 92.0 & 91.3 & 88.7 & $+3.3$ [-1.3, 8.7] & $+2$ & $-2$ \\
 & MMLU-Pro & 63.3 & 64.7 & 60.0 & $+3.3$ [-4.0, 10.7] & $+11$ & $-11$ \\
Qwen3-8B & GSM8K & 92.7 & 94.0 & 94.7 & $-2.0$ [-5.3, 1.3] & $+1$ & $+2$ \\
 & MMLU-Pro & 60.7 & 60.0 & 58.0 & $+2.7$ [-4.7, 9.3] & $+6$ & $-4$ \\
Qwen3-1.7B & GSM8K & 79.3 & 75.3 & 74.0 & $+5.3$ [-2.0, 13.3] & $+3$ & $+5$ \\
 & MMLU-Pro & 48.0 & 47.3 & 47.3 & $+0.7$ [-7.3, 8.7] & $+6$ & $-5$ \\
Qwen3-4B & GSM8K & 96.0 & 96.7 & 93.3 & $+2.7$ [-1.3, 6.7] & $+0$ & $+0$ \\
 & MMLU-Pro & 64.0 & 60.7 & 64.7 & $-0.7$ [-6.0, 4.7] & $+2$ & $-1$ \\
\bottomrule
\end{tabular}

\end{table*}

\begin{table*}[t]
\begin{minipage}[t]{0.46\textwidth}\centering
\caption{Plain-temperature accuracy drop from $T{=}0.7$ to $1.3$ (pp, point estimates) at each
quantization level of the main grid (150 generations per cell).}
\label{tab:quant-drops}
\footnotesize\setlength{\tabcolsep}{4pt}\resizebox{\linewidth}{!}{\begin{tabular}{llrrrr}
\toprule
Model & Task & Q8\_0 & Q6\_K & Q4\_K\_M & Q3\_K\_M \\
\midrule
Llama-3.1-8B & GSM8K & $+30.0$ & $+32.0$ & $+36.0$ & $+38.0$ \\
 & MMLU-Pro & $+38.0$ & $+46.7$ & $+38.0$ & $+40.7$ \\
Mistral-7B & GSM8K & $-0.7$ & $-0.7$ & $+3.3$ & $+12.0$ \\
 & MMLU-Pro & $+9.3$ & $+5.3$ & $+10.7$ & $+6.7$ \\
Qwen2.5-7B & GSM8K & $+4.0$ & $+2.7$ & $+2.7$ & $+6.7$ \\
 & MMLU-Pro & $+7.3$ & $+6.0$ & $+4.0$ & $+2.0$ \\
Gemma-3-12B & GSM8K & $+3.3$ & $+2.7$ & $+0.0$ & $+2.0$ \\
 & MMLU-Pro & $+3.3$ & $+5.3$ & $+2.7$ & $+3.3$ \\
Qwen3-8B & GSM8K & $-2.0$ & $+1.3$ & $+2.0$ & $+0.7$ \\
 & MMLU-Pro & $+2.7$ & $-0.7$ & $+0.0$ & $+2.0$ \\
Qwen3-4B & GSM8K & $+2.7$ & $+0.7$ & $+6.0$ & $+2.7$ \\
 & MMLU-Pro & $-0.7$ & $+0.0$ & $+2.0$ & $+4.0$ \\
Qwen3-1.7B & GSM8K & $+5.3$ & $-6.0$ & $-0.7$ & $+0.0$ \\
 & MMLU-Pro & $+0.7$ & $+6.0$ & $-8.0$ & $+4.0$ \\
\bottomrule
\end{tabular}
}
\end{minipage}\hfill
\begin{minipage}[t]{0.5\textwidth}\centering
\caption{Stratified MMLU-Pro subset against the fixed business subset: plain-temperature
accuracy at Q8\_0 and the drop with its interval.}
\label{tab:strat}
\footnotesize\setlength{\tabcolsep}{4pt}\resizebox{\linewidth}{!}{\begin{tabular}{llccl}
\toprule
Model & Subset & acc.\ 0.7 & acc.\ 1.3 & drop [95\% CI] \\
\midrule
Llama-3.1-8B & business (head) & 48.0 & 10.0 & $+38.0$ [28.0, 48.0] \\
 & stratified & 44.7 & 1.3 & $+43.3$ [31.3, 55.3] \\
Qwen2.5-7B & business (head) & 65.3 & 58.0 & $+7.3$ [-1.3, 16.0] \\
 & stratified & 54.7 & 48.7 & $+6.0$ [-2.0, 14.7] \\
Gemma-3-12B & business (head) & 63.3 & 60.0 & $+3.3$ [-4.0, 10.7] \\
 & stratified & 64.0 & 56.7 & $+7.3$ [2.0, 12.7] \\
\bottomrule
\end{tabular}
}
\end{minipage}
\end{table*}

\section{Temperature sweep}
\label{app:ladder}

The sweep of Section~\ref{sec:recovery} is a separate run from the main grid.
Table~\ref{tab:ladder} gives the plain-temperature drop at each sweep temperature against the
model's own accuracy at 0.7 in the main grid, with intervals. The 1.3 step of the sweep differs
from the main-grid value of Table~\ref{tab:temp-full} by amounts within the intervals
(Llama-3.1 on MMLU-Pro: $+44.7$ in the sweep against $+38.0$ in the grid).

\begin{table*}[p]\centering
\caption{Plain-temperature accuracy drop at each sweep temperature against the model's own
$T{=}0.7$ accuracy in the main grid (pp, 95\% item-clustered intervals; Q8\_0).}
\label{tab:ladder}
\footnotesize\setlength{\tabcolsep}{6pt}\begin{tabular}{lclll}
\toprule
Task & $T$ & Llama-3.1-8B & Qwen2.5-7B & Gemma-3-12B \\
\midrule
GSM8K & 1.3 & $+32.7$ [24.7, 41.3] & $+4.0$ [-0.7, 9.3] & $+2.7$ [-2.0, 8.0] \\
 & 1.5 & $+68.0$ [58.7, 76.7] & $+24.0$ [15.3, 32.7] & $+14.7$ [8.7, 22.0] \\
 & 1.7 & $+76.0$ [66.0, 85.3] & $+70.0$ [61.3, 77.3] & $+42.7$ [34.7, 50.7] \\
 & 2.0 & $+77.3$ [67.3, 86.7] & $+90.7$ [82.0, 97.3] & $+84.0$ [76.0, 90.7] \\
\midrule
MMLU-Pro & 1.3 & $+44.7$ [34.6, 55.3] & $+8.0$ [0.0, 17.3] & $+3.3$ [-2.7, 10.0] \\
 & 1.5 & $+46.0$ [35.3, 56.7] & $+40.0$ [28.6, 51.3] & $+39.3$ [29.3, 50.0] \\
 & 1.7 & $+48.0$ [36.7, 59.3] & $+61.3$ [49.3, 73.3] & $+57.3$ [46.0, 69.3] \\
 & 2.0 & $+48.0$ [36.7, 59.3] & $+65.3$ [53.3, 77.3] & $+63.3$ [51.3, 75.3] \\
\bottomrule
\end{tabular}

\end{table*}

\section{Sampler results in full}
\label{app:sampler}

Tables~\ref{tab:grid-gsm8k} and \ref{tab:grid-mmlu} give the accuracy of every decoding
configuration at every temperature for the ten models with a sampler grid, with the main-grid
models pooled over their quantization levels, Qwen3-4B's 16-bit run included, and the three
flagged models at Q8\_0.
Tables~\ref{tab:contrasts-gsm8k} and \ref{tab:contrasts-mmlu} give every paired difference
from plain temperature sampling, including the two configurations that apply truncation before
temperature scaling. An interval endpoint within $10^{-9}$ of zero counts as touching zero.
The upper bound of Table~\ref{tab:bound} pools the four quantization levels of the main grid
and is also given at Q8\_0 alone. A generalized estimating equation over the same records
(identity link, item clusters, exchangeable working correlation) gives the same point
estimates and intervals that differ by at most 0.6 points (median 0.1).

\textbf{Order of temperature and truncation.} Applying top-$p$ or min-$p$ before temperature
scaling, the llama.cpp default order, changes accuracy by at most 4 points on the seven
main-grid models at temperatures 0.7 and 1.0. The largest gain over plain temperature sampling
anywhere at those temperatures belongs to this order: top-$p$ applied first on Hermes-3 at 0.7
on MMLU-Pro gains $+12.0$ [4.7, 19.3], the largest of the 72 flagged-model comparisons at those
temperatures; the other 71 stay within 8 points. On Llama-3.1 at 1.3, truncating first keeps 22 more points
under top-$p$ on MMLU-Pro, because at that temperature the sampler sees a sharper distribution
and removes more of the tail.

\begin{figure*}[p]
\centering
\includegraphics[width=0.88\columnwidth]{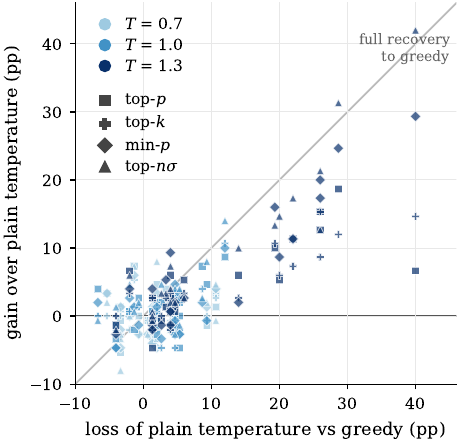}
\caption{Gain of each truncation sampler over plain temperature sampling against the loss of
plain temperature sampling relative to greedy decoding: 240 points, four samplers in each of
the 60 combinations of model, task, and temperature with a sampler grid (Q8\_0). Both axes
subtract the same plain-temperature accuracy. The diagonal marks full recovery to greedy
accuracy.}
\label{fig:gainloss}
\end{figure*}

\textbf{Loss and gain across cells.} Figure~\ref{fig:gainloss} plots, for each of the 60
combinations of model, task, and temperature with a sampler grid, the gain of each truncation
sampler over plain temperature sampling against the loss of plain temperature sampling relative
to greedy decoding. Larger losses accompany larger gains (Pearson $r{=}0.89$ over the 60
best-sampler points and 0.78 over all 240 points). The two quantities share the
plain-temperature baseline and the points are not independent observations, so the
association is descriptive.

\begin{table*}[p]\centering
\caption{GSM8K accuracy (\%) by configuration and temperature (main-grid models pooled over
quantization levels; the three flagged models at Q8\_0).}
\label{tab:grid-gsm8k}
\footnotesize\setlength{\tabcolsep}{4pt}\resizebox{\textwidth}{!}{\begin{tabular}{llccc@{\hspace{14pt}}llccc}
\toprule
Model & Configuration & 0.7 & 1.0 & 1.3 & Model & Configuration & 0.7 & 1.0 & 1.3 \\
\midrule
Llama-3.1-8B & greedy & 78.0 & -- & -- & Qwen3-4B & greedy & 95.2 & -- & -- \\
 & temperature & 78.2 & 74.8 & 44.2 &  & temperature & 95.2 & 94.0 & 93.1 \\
 & top-$p$ & 81.5 & 78.0 & 63.8 &  & top-$p$ & 95.5 & 95.5 & 93.5 \\
 & top-$k$ & 79.8 & 75.2 & 58.7 &  & top-$k$ & 94.0 & 94.5 & 94.1 \\
 & min-$p$ & 81.2 & 78.0 & 72.0 &  & min-$p$ & 95.1 & 94.7 & 92.9 \\
 & top-$n\sigma$ & 78.8 & 80.7 & 80.8 &  & top-$n\sigma$ & 94.8 & 93.9 & 93.9 \\
 & top-$p$ (temp.\ last) & 80.8 & 78.7 & 73.0 &  & top-$p$ (temp.\ last) & 94.8 & 95.2 & 94.3 \\
 & min-$p$ (temp.\ last) & 78.7 & 81.3 & 76.2 &  & min-$p$ (temp.\ last) & 94.5 & 93.9 & 93.9 \\
\midrule
Mistral-7B & greedy & 31.0 & -- & -- & Qwen3-1.7B & greedy & 75.5 & -- & -- \\
 & temperature & 32.7 & 30.0 & 29.2 &  & temperature & 73.3 & 73.7 & 73.7 \\
 & top-$p$ & 28.2 & 31.0 & 34.2 &  & top-$p$ & 73.8 & 75.0 & 72.7 \\
 & top-$k$ & 31.8 & 31.7 & 29.8 &  & top-$k$ & 72.8 & 72.7 & 73.3 \\
 & min-$p$ & 32.3 & 28.3 & 28.8 &  & min-$p$ & 72.2 & 71.7 & 74.0 \\
 & top-$n\sigma$ & 29.5 & 30.7 & 30.2 &  & top-$n\sigma$ & 74.3 & 73.5 & 72.3 \\
 & top-$p$ (temp.\ last) & 29.5 & 31.2 & 29.5 &  & top-$p$ (temp.\ last) & 73.0 & 72.0 & 71.3 \\
 & min-$p$ (temp.\ last) & 29.5 & 32.0 & 30.5 &  & min-$p$ (temp.\ last) & 74.7 & 73.3 & 74.5 \\
\midrule
Qwen2.5-7B & greedy & 93.5 & -- & -- & Hermes-3-8B & greedy & 82.0 & -- & -- \\
 & temperature & 91.5 & 91.7 & 87.5 &  & temperature & 71.3 & 77.3 & 60.0 \\
 & top-$p$ & 92.2 & 91.2 & 90.3 &  & top-$p$ & 74.7 & 80.0 & 71.3 \\
 & top-$k$ & 90.2 & 88.5 & 88.7 &  & top-$k$ & 74.0 & 72.7 & 67.3 \\
 & min-$p$ & 90.8 & 90.5 & 91.0 &  & min-$p$ & 75.3 & 76.0 & 71.3 \\
 & top-$n\sigma$ & 92.7 & 91.2 & 90.7 &  & top-$n\sigma$ & 70.7 & 74.7 & 77.3 \\
 & top-$p$ (temp.\ last) & 90.8 & 91.3 & 90.3 &  & top-$p$ (temp.\ last) & 72.0 & 73.3 & 77.3 \\
 & min-$p$ (temp.\ last) & 92.0 & 90.7 & 91.3 &  & min-$p$ (temp.\ last) & 75.3 & 77.3 & 74.7 \\
\midrule
Gemma-3-12B & greedy & 92.0 & -- & -- & Qwen3.5-9B & greedy & 78.0 & -- & -- \\
 & temperature & 91.0 & 90.2 & 89.0 &  & temperature & 81.3 & 80.0 & 74.0 \\
 & top-$p$ & 91.5 & 90.3 & 90.7 &  & top-$p$ & 80.7 & 80.7 & 80.0 \\
 & top-$k$ & 91.8 & 91.7 & 90.7 &  & top-$k$ & 81.3 & 80.7 & 77.3 \\
 & min-$p$ & 91.5 & 91.3 & 92.3 &  & min-$p$ & 82.7 & 82.7 & 83.3 \\
 & top-$n\sigma$ & 91.3 & 91.5 & 91.0 &  & top-$n\sigma$ & 78.7 & 84.7 & 81.3 \\
 & top-$p$ (temp.\ last) & 90.2 & 91.8 & 91.5 &  & top-$p$ (temp.\ last) & 82.7 & 80.7 & 82.0 \\
 & min-$p$ (temp.\ last) & 91.2 & 90.8 & 90.3 &  & min-$p$ (temp.\ last) & 82.0 & 78.0 & 79.3 \\
\midrule
Qwen3-8B & greedy & 94.5 & -- & -- & OLMo-3-7B & greedy & 90.0 & -- & -- \\
 & temperature & 95.0 & 94.2 & 94.5 &  & temperature & 88.7 & 81.3 & 76.0 \\
 & top-$p$ & 94.0 & 94.3 & 93.7 &  & top-$p$ & 84.7 & 88.7 & 82.0 \\
 & top-$k$ & 94.8 & 94.8 & 93.5 &  & top-$k$ & 84.7 & 85.3 & 78.7 \\
 & min-$p$ & 94.8 & 95.3 & 94.7 &  & min-$p$ & 84.7 & 83.3 & 78.0 \\
 & top-$n\sigma$ & 96.3 & 94.2 & 94.3 &  & top-$n\sigma$ & 86.7 & 84.0 & 86.0 \\
 & top-$p$ (temp.\ last) & 93.8 & 94.0 & 94.0 &  & top-$p$ (temp.\ last) & 84.0 & 85.3 & 81.3 \\
 & min-$p$ (temp.\ last) & 96.2 & 94.7 & 95.3 &  & min-$p$ (temp.\ last) & 83.3 & 84.7 & 86.0 \\
\bottomrule
\end{tabular}
}
\end{table*}
\begin{table*}[p]\centering
\caption{MMLU-Pro accuracy (\%) by configuration and temperature (main-grid models pooled over
quantization levels; the three flagged models at Q8\_0).}
\label{tab:grid-mmlu}
\footnotesize\setlength{\tabcolsep}{4pt}\resizebox{\textwidth}{!}{\begin{tabular}{llccc@{\hspace{14pt}}llccc}
\toprule
Model & Configuration & 0.7 & 1.0 & 1.3 & Model & Configuration & 0.7 & 1.0 & 1.3 \\
\midrule
Llama-3.1-8B & greedy & 54.0 & -- & -- & Qwen3-4B & greedy & 61.6 & -- & -- \\
 & temperature & 47.8 & 39.7 & 7.0 &  & temperature & 61.9 & 62.5 & 61.2 \\
 & top-$p$ & 51.3 & 46.0 & 16.0 &  & top-$p$ & 62.8 & 61.3 & 62.8 \\
 & top-$k$ & 51.2 & 43.3 & 26.8 &  & top-$k$ & 63.5 & 61.1 & 61.2 \\
 & min-$p$ & 51.2 & 50.8 & 39.5 &  & min-$p$ & 60.9 & 62.7 & 62.9 \\
 & top-$n\sigma$ & 51.0 & 50.5 & 48.8 &  & top-$n\sigma$ & 62.4 & 62.4 & 61.5 \\
 & top-$p$ (temp.\ last) & 50.3 & 46.7 & 37.5 &  & top-$p$ (temp.\ last) & 62.9 & 62.7 & 61.7 \\
 & min-$p$ (temp.\ last) & 51.2 & 50.2 & 43.5 &  & min-$p$ (temp.\ last) & 61.3 & 63.6 & 61.5 \\
\midrule
Mistral-7B & greedy & 35.5 & -- & -- & Qwen3-1.7B & greedy & 45.5 & -- & -- \\
 & temperature & 30.5 & 30.2 & 22.5 &  & temperature & 46.0 & 45.7 & 45.3 \\
 & top-$p$ & 30.2 & 29.5 & 25.3 &  & top-$p$ & 47.2 & 47.0 & 47.2 \\
 & top-$k$ & 32.5 & 32.3 & 28.3 &  & top-$k$ & 47.5 & 46.5 & 45.0 \\
 & min-$p$ & 33.8 & 30.7 & 31.0 &  & min-$p$ & 47.0 & 47.7 & 46.2 \\
 & top-$n\sigma$ & 32.3 & 31.5 & 31.5 &  & top-$n\sigma$ & 46.8 & 47.2 & 46.2 \\
 & top-$p$ (temp.\ last) & 29.2 & 30.2 & 32.5 &  & top-$p$ (temp.\ last) & 47.7 & 46.8 & 43.5 \\
 & min-$p$ (temp.\ last) & 33.2 & 33.3 & 28.7 &  & min-$p$ (temp.\ last) & 45.7 & 46.2 & 49.0 \\
\midrule
Qwen2.5-7B & greedy & 64.0 & -- & -- & Hermes-3-8B & greedy & 52.0 & -- & -- \\
 & temperature & 62.5 & 61.3 & 57.7 &  & temperature & 42.7 & 42.7 & 26.0 \\
 & top-$p$ & 65.5 & 62.8 & 62.7 &  & top-$p$ & 41.3 & 47.3 & 38.7 \\
 & top-$k$ & 62.5 & 61.7 & 57.0 &  & top-$k$ & 42.7 & 42.0 & 34.7 \\
 & min-$p$ & 61.3 & 63.5 & 59.7 &  & min-$p$ & 44.0 & 42.0 & 46.0 \\
 & top-$n\sigma$ & 62.0 & 63.5 & 61.0 &  & top-$n\sigma$ & 44.7 & 50.7 & 38.7 \\
 & top-$p$ (temp.\ last) & 63.7 & 62.7 & 61.3 &  & top-$p$ (temp.\ last) & 54.7 & 46.7 & 43.3 \\
 & min-$p$ (temp.\ last) & 64.2 & 63.0 & 62.3 &  & min-$p$ (temp.\ last) & 44.0 & 44.7 & 44.0 \\
\midrule
Gemma-3-12B & greedy & 61.0 & -- & -- & Qwen3.5-9B & greedy & 64.0 & -- & -- \\
 & temperature & 63.7 & 66.7 & 60.0 &  & temperature & 60.0 & 58.7 & 38.0 \\
 & top-$p$ & 65.7 & 66.7 & 63.8 &  & top-$p$ & 62.0 & 60.7 & 53.3 \\
 & top-$k$ & 64.3 & 62.8 & 63.7 &  & top-$k$ & 59.3 & 56.7 & 53.3 \\
 & min-$p$ & 65.3 & 67.0 & 65.5 &  & min-$p$ & 61.3 & 56.0 & 55.3 \\
 & top-$n\sigma$ & 65.2 & 65.8 & 65.8 &  & top-$n\sigma$ & 58.0 & 61.3 & 59.3 \\
 & top-$p$ (temp.\ last) & 65.8 & 65.0 & 65.2 &  & top-$p$ (temp.\ last) & 56.7 & 60.7 & 54.0 \\
 & min-$p$ (temp.\ last) & 65.3 & 63.7 & 63.8 &  & min-$p$ (temp.\ last) & 60.0 & 57.3 & 57.3 \\
\midrule
Qwen3-8B & greedy & 60.5 & -- & -- & OLMo-3-7B & greedy & 46.0 & -- & -- \\
 & temperature & 60.5 & 59.0 & 59.5 &  & temperature & 42.0 & 41.3 & 26.7 \\
 & top-$p$ & 60.5 & 61.3 & 60.5 &  & top-$p$ & 43.3 & 39.3 & 36.7 \\
 & top-$k$ & 60.2 & 61.3 & 57.0 &  & top-$k$ & 40.0 & 40.7 & 37.3 \\
 & min-$p$ & 61.0 & 59.2 & 59.7 &  & min-$p$ & 43.3 & 40.7 & 42.7 \\
 & top-$n\sigma$ & 60.0 & 58.8 & 58.7 &  & top-$n\sigma$ & 41.3 & 42.7 & 40.0 \\
 & top-$p$ (temp.\ last) & 60.7 & 59.5 & 60.5 &  & top-$p$ (temp.\ last) & 43.3 & 44.0 & 44.7 \\
 & min-$p$ (temp.\ last) & 61.3 & 60.7 & 59.3 &  & min-$p$ (temp.\ last) & 42.0 & 44.7 & 40.0 \\
\bottomrule
\end{tabular}
}
\end{table*}
\begin{table*}[p]\centering
\caption{GSM8K: paired accuracy difference of each configuration against plain temperature
sampling on the same items (pp, 95\% item-clustered bootstrap intervals; main-grid models
pooled over quantization, flagged models at Q8\_0). Bold: interval above zero.}
\label{tab:contrasts-gsm8k}
\scriptsize\setlength{\tabcolsep}{3pt}\begin{tabular}{lcllllll}
\toprule
Model & $T$ & top-$p$ & top-$k$ & min-$p$ & top-$n\sigma$ & top-$p$ (temp.\ last) & min-$p$ (temp.\ last) \\
\midrule
Llama-3.1-8B & 0.7 & $\mathbf{+3.3}$ [0.3, 6.8] & $+1.7$ [-1.0, 4.5] & $+3.0$ [-0.8, 7.0] & $+0.7$ [-2.7, 4.0] & $+2.7$ [-0.5, 6.3] & $+0.5$ [-2.3, 3.3] \\
 & 1.0 & $+3.2$ [-0.3, 6.8] & $+0.3$ [-3.5, 4.5] & $+3.2$ [-0.8, 7.3] & $\mathbf{+5.8}$ [2.2, 9.5] & $\mathbf{+3.8}$ [0.2, 8.0] & $\mathbf{+6.5}$ [2.8, 10.5] \\
 & 1.3 & $\mathbf{+19.7}$ [15.0, 24.3] & $\mathbf{+14.5}$ [10.0, 18.8] & $\mathbf{+27.8}$ [22.7, 33.2] & $\mathbf{+36.7}$ [30.3, 43.2] & $\mathbf{+28.8}$ [23.7, 34.2] & $\mathbf{+32.0}$ [26.0, 38.3] \\
\midrule
Mistral-7B & 0.7 & $-4.5$ [-7.7, -1.5] & $-0.8$ [-5.8, 3.7] & $-0.3$ [-3.8, 3.3] & $-3.2$ [-7.7, 1.0] & $-3.2$ [-8.0, 1.5] & $-3.2$ [-7.2, 0.7] \\
 & 1.0 & $+1.0$ [-2.5, 4.7] & $+1.7$ [-1.5, 4.7] & $-1.7$ [-4.8, 1.7] & $+0.7$ [-2.8, 4.2] & $+1.2$ [-3.0, 5.5] & $+2.0$ [-1.8, 6.0] \\
 & 1.3 & $\mathbf{+5.0}$ [0.7, 9.2] & $+0.7$ [-4.3, 5.5] & $-0.3$ [-4.0, 3.5] & $+1.0$ [-2.8, 5.0] & $+0.3$ [-4.8, 5.0] & $+1.3$ [-2.3, 5.0] \\
\midrule
Qwen2.5-7B & 0.7 & $+0.7$ [-1.3, 2.8] & $-1.3$ [-3.8, 0.7] & $-0.7$ [-2.0, 0.7] & $+1.2$ [-0.7, 3.2] & $-0.7$ [-2.8, 1.3] & $+0.5$ [-1.0, 2.2] \\
 & 1.0 & $-0.5$ [-3.0, 1.7] & $-3.2$ [-6.0, -0.5] & $-1.2$ [-3.5, 1.2] & $-0.5$ [-2.8, 1.5] & $-0.3$ [-3.3, 2.5] & $-1.0$ [-3.7, 1.5] \\
 & 1.3 & $\mathbf{+2.8}$ [0.7, 5.2] & $+1.2$ [-0.7, 3.3] & $\mathbf{+3.5}$ [0.8, 6.5] & $\mathbf{+3.2}$ [1.2, 5.5] & $\mathbf{+2.8}$ [0.5, 5.5] & $\mathbf{+3.8}$ [1.5, 6.7] \\
\midrule
Gemma-3-12B & 0.7 & $+0.5$ [-1.3, 2.5] & $+0.8$ [-1.0, 3.0] & $+0.5$ [-1.3, 2.5] & $+0.3$ [-1.5, 2.5] & $-0.8$ [-4.0, 2.2] & $+0.2$ [-2.0, 2.5] \\
 & 1.0 & $+0.2$ [-3.0, 3.0] & $+1.5$ [-0.5, 3.7] & $+1.2$ [-1.2, 3.7] & $+1.3$ [-1.0, 3.7] & $+1.7$ [-1.0, 4.7] & $+0.7$ [-1.5, 2.8] \\
 & 1.3 & $+1.7$ [-0.2, 3.8] & $+1.7$ [-0.7, 4.2] & $\mathbf{+3.3}$ [0.5, 6.7] & $+2.0$ [0.0, 4.5] & $\mathbf{+2.5}$ [0.3, 4.8] & $+1.3$ [-0.8, 3.5] \\
\midrule
Qwen3-8B & 0.7 & $-1.0$ [-3.3, 0.8] & $-0.2$ [-1.7, 1.0] & $-0.2$ [-1.7, 1.2] & $+1.3$ [-0.8, 3.5] & $-1.2$ [-3.5, 0.8] & $+1.2$ [-0.8, 3.5] \\
 & 1.0 & $+0.2$ [-1.7, 2.0] & $+0.7$ [-2.0, 3.3] & $+1.2$ [-1.0, 3.3] & $+0.0$ [-1.8, 1.7] & $-0.2$ [-2.0, 2.0] & $+0.5$ [-1.8, 2.8] \\
 & 1.3 & $-0.8$ [-2.5, 0.7] & $-1.0$ [-2.7, 0.5] & $+0.2$ [-1.8, 2.3] & $-0.2$ [-1.7, 1.7] & $-0.5$ [-2.2, 1.0] & $+0.8$ [-0.3, 2.2] \\
\midrule
Qwen3-4B & 0.7 & $+0.3$ [-0.9, 1.7] & $-1.2$ [-2.9, 0.8] & $-0.1$ [-1.3, 1.3] & $-0.4$ [-1.7, 1.1] & $-0.4$ [-2.0, 1.6] & $-0.7$ [-2.5, 1.6] \\
 & 1.0 & $+1.5$ [0.0, 3.1] & $+0.5$ [-0.5, 1.7] & $+0.7$ [-1.1, 2.7] & $-0.1$ [-1.5, 1.1] & $+1.2$ [0.0, 2.4] & $-0.1$ [-1.6, 1.3] \\
 & 1.3 & $+0.4$ [-1.2, 2.0] & $+1.1$ [-0.7, 2.8] & $-0.1$ [-1.9, 1.6] & $+0.8$ [-1.1, 2.7] & $+1.2$ [-0.5, 3.1] & $+0.8$ [-0.8, 2.5] \\
\midrule
Qwen3-1.7B & 0.7 & $+0.5$ [-2.8, 3.8] & $-0.5$ [-4.2, 2.8] & $-1.2$ [-4.7, 2.2] & $+1.0$ [-2.2, 4.3] & $-0.3$ [-3.0, 2.2] & $+1.3$ [-1.3, 4.3] \\
 & 1.0 & $+1.3$ [-2.0, 4.5] & $-1.0$ [-4.0, 2.0] & $-2.0$ [-4.3, 0.0] & $-0.2$ [-2.7, 2.3] & $-1.7$ [-4.0, 0.5] & $-0.3$ [-3.2, 2.7] \\
 & 1.3 & $-1.0$ [-4.5, 2.3] & $-0.3$ [-3.5, 2.7] & $+0.3$ [-2.8, 3.5] & $-1.3$ [-4.2, 1.2] & $-2.3$ [-6.0, 1.2] & $+0.8$ [-2.2, 3.7] \\
\midrule
Hermes-3-8B & 0.7 & $+3.3$ [-3.3, 10.0] & $+2.7$ [-3.3, 8.7] & $+4.0$ [-2.0, 10.0] & $-0.7$ [-6.0, 4.7] & $+0.7$ [-6.0, 7.3] & $+4.0$ [-2.7, 10.0] \\
 & 1.0 & $+2.7$ [-3.3, 8.0] & $-4.7$ [-10.7, 1.3] & $-1.3$ [-6.7, 4.0] & $-2.7$ [-9.3, 3.3] & $-4.0$ [-10.0, 2.0] & $+0.0$ [-7.3, 7.3] \\
 & 1.3 & $\mathbf{+11.3}$ [2.7, 19.3] & $+7.3$ [-0.7, 14.7] & $\mathbf{+11.3}$ [3.3, 19.3] & $\mathbf{+17.3}$ [10.7, 24.0] & $\mathbf{+17.3}$ [8.7, 25.3] & $\mathbf{+14.7}$ [7.3, 22.0] \\
\midrule
Qwen3.5-9B & 0.7 & $-0.7$ [-4.7, 3.3] & $+0.0$ [-5.3, 5.3] & $+1.3$ [-3.3, 5.3] & $-2.7$ [-8.0, 2.0] & $+1.3$ [-3.3, 5.3] & $+0.7$ [-2.7, 4.7] \\
 & 1.0 & $+0.7$ [-5.3, 6.7] & $+0.7$ [-6.0, 7.3] & $+2.7$ [-1.3, 7.3] & $+4.7$ [-1.3, 11.3] & $+0.7$ [-4.7, 6.7] & $-2.0$ [-6.0, 1.3] \\
 & 1.3 & $\mathbf{+6.0}$ [1.3, 11.3] & $+3.3$ [-4.0, 10.7] & $\mathbf{+9.3}$ [0.7, 18.0] & $\mathbf{+7.3}$ [1.3, 13.3] & $\mathbf{+8.0}$ [2.0, 14.7] & $+5.3$ [-1.3, 12.0] \\
\midrule
OLMo-3-7B & 0.7 & $-4.0$ [-8.7, 0.7] & $-4.0$ [-9.3, 0.0] & $-4.0$ [-9.3, 0.7] & $-2.0$ [-5.3, 1.3] & $-4.7$ [-10.7, 0.7] & $-5.3$ [-10.7, -1.3] \\
 & 1.0 & $\mathbf{+7.3}$ [3.3, 12.0] & $+4.0$ [0.0, 8.7] & $+2.0$ [-2.7, 6.7] & $+2.7$ [-3.3, 8.7] & $+4.0$ [0.0, 8.7] & $+3.3$ [-0.7, 8.0] \\
 & 1.3 & $\mathbf{+6.0}$ [1.3, 10.7] & $+2.7$ [-3.3, 9.3] & $+2.0$ [-4.0, 9.3] & $\mathbf{+10.0}$ [3.3, 17.3] & $+5.3$ [-0.7, 11.3] & $\mathbf{+10.0}$ [3.3, 17.3] \\
\bottomrule
\end{tabular}

\end{table*}
\begin{table*}[p]\centering
\caption{MMLU-Pro: paired accuracy difference of each configuration against plain temperature
sampling on the same items, as in Table~\ref{tab:contrasts-gsm8k}.}
\label{tab:contrasts-mmlu}
\scriptsize\setlength{\tabcolsep}{3pt}\begin{tabular}{lcllllll}
\toprule
Model & $T$ & top-$p$ & top-$k$ & min-$p$ & top-$n\sigma$ & top-$p$ (temp.\ last) & min-$p$ (temp.\ last) \\
\midrule
Llama-3.1-8B & 0.7 & $\mathbf{+3.5}$ [0.3, 7.0] & $+3.3$ [-0.5, 7.5] & $+3.3$ [-0.7, 8.0] & $+3.2$ [-0.2, 6.7] & $+2.5$ [-0.7, 6.2] & $\mathbf{+3.3}$ [0.3, 6.7] \\
 & 1.0 & $\mathbf{+6.3}$ [1.8, 11.3] & $+3.7$ [-1.3, 8.7] & $\mathbf{+11.2}$ [6.7, 16.0] & $\mathbf{+10.8}$ [6.0, 16.2] & $\mathbf{+7.0}$ [3.0, 11.2] & $\mathbf{+10.5}$ [6.3, 14.7] \\
 & 1.3 & $\mathbf{+9.0}$ [5.2, 13.3] & $\mathbf{+19.8}$ [13.8, 26.7] & $\mathbf{+32.5}$ [24.7, 41.0] & $\mathbf{+41.8}$ [32.3, 52.2] & $\mathbf{+30.5}$ [23.2, 38.7] & $\mathbf{+36.5}$ [28.0, 45.5] \\
\midrule
Mistral-7B & 0.7 & $-0.3$ [-4.5, 4.0] & $+2.0$ [-2.3, 6.5] & $+3.3$ [-0.5, 7.0] & $+1.8$ [-2.2, 6.0] & $-1.3$ [-4.8, 2.2] & $+2.7$ [-1.7, 7.3] \\
 & 1.0 & $-0.7$ [-4.5, 3.0] & $+2.2$ [-1.2, 5.5] & $+0.5$ [-2.8, 3.8] & $+1.3$ [-3.2, 5.8] & $+0.0$ [-4.0, 4.5] & $+3.2$ [-0.8, 7.3] \\
 & 1.3 & $+2.8$ [-1.0, 7.0] & $\mathbf{+5.8}$ [2.7, 9.3] & $\mathbf{+8.5}$ [4.5, 12.5] & $\mathbf{+9.0}$ [4.8, 13.5] & $\mathbf{+10.0}$ [5.3, 15.3] & $\mathbf{+6.2}$ [2.0, 10.3] \\
\midrule
Qwen2.5-7B & 0.7 & $\mathbf{+3.0}$ [0.5, 5.8] & $+0.0$ [-2.7, 2.7] & $-1.2$ [-4.5, 2.2] & $-0.5$ [-3.7, 2.5] & $+1.2$ [-1.2, 3.7] & $+1.7$ [-1.2, 4.7] \\
 & 1.0 & $+1.5$ [-1.8, 5.0] & $+0.3$ [-2.0, 2.7] & $+2.2$ [-0.8, 5.3] & $+2.2$ [-0.8, 5.3] & $+1.3$ [-1.0, 3.8] & $+1.7$ [-1.2, 4.8] \\
 & 1.3 & $\mathbf{+5.0}$ [1.7, 8.3] & $-0.7$ [-3.8, 2.3] & $+2.0$ [-1.7, 6.5] & $+3.3$ [-0.2, 7.0] & $\mathbf{+3.7}$ [0.2, 7.2] & $\mathbf{+4.7}$ [1.0, 8.8] \\
\midrule
Gemma-3-12B & 0.7 & $+2.0$ [-0.3, 4.5] & $+0.7$ [-2.0, 3.5] & $+1.7$ [-0.8, 4.3] & $+1.5$ [-1.2, 4.3] & $+2.2$ [-1.0, 5.7] & $+1.7$ [-2.0, 5.7] \\
 & 1.0 & $+0.0$ [-2.8, 2.7] & $-3.8$ [-7.2, -1.0] & $+0.3$ [-2.2, 2.7] & $-0.8$ [-3.0, 1.3] & $-1.7$ [-4.7, 1.3] & $-3.0$ [-6.2, -0.2] \\
 & 1.3 & $\mathbf{+3.8}$ [0.2, 8.3] & $\mathbf{+3.7}$ [0.3, 7.5] & $\mathbf{+5.5}$ [2.7, 8.8] & $\mathbf{+5.8}$ [2.3, 9.8] & $\mathbf{+5.2}$ [2.3, 8.3] & $\mathbf{+3.8}$ [0.5, 7.5] \\
\midrule
Qwen3-8B & 0.7 & $+0.0$ [-2.8, 3.0] & $-0.3$ [-2.7, 2.0] & $+0.5$ [-2.3, 3.7] & $-0.5$ [-2.8, 1.8] & $+0.2$ [-2.7, 2.8] & $+0.8$ [-2.2, 3.8] \\
 & 1.0 & $+2.3$ [-0.5, 5.2] & $+2.3$ [-0.7, 5.3] & $+0.2$ [-3.2, 3.3] & $-0.2$ [-3.2, 2.7] & $+0.5$ [-2.3, 3.2] & $+1.7$ [-1.2, 4.5] \\
 & 1.3 & $+1.0$ [-1.8, 3.5] & $-2.5$ [-5.7, 0.7] & $+0.2$ [-3.0, 3.5] & $-0.8$ [-4.2, 2.3] & $+1.0$ [-1.8, 3.8] & $-0.2$ [-3.5, 3.0] \\
\midrule
Qwen3-4B & 0.7 & $+0.9$ [-1.5, 3.1] & $+1.6$ [-0.5, 3.7] & $-0.9$ [-3.5, 1.6] & $+0.5$ [-1.5, 2.5] & $+1.1$ [-1.1, 3.2] & $-0.5$ [-3.1, 2.0] \\
 & 1.0 & $-1.2$ [-3.2, 0.9] & $-1.5$ [-3.5, 0.4] & $+0.1$ [-2.7, 3.2] & $-0.1$ [-2.7, 2.4] & $+0.1$ [-2.7, 3.2] & $+1.1$ [-1.5, 3.7] \\
 & 1.3 & $+1.6$ [-1.2, 4.3] & $+0.0$ [-2.7, 2.5] & $+1.7$ [-0.7, 4.1] & $+0.3$ [-2.3, 2.8] & $+0.5$ [-2.3, 3.2] & $+0.3$ [-2.1, 2.8] \\
\midrule
Qwen3-1.7B & 0.7 & $+1.2$ [-2.0, 4.2] & $+1.5$ [-1.7, 4.7] & $+1.0$ [-2.5, 4.7] & $+0.8$ [-2.5, 4.2] & $+1.7$ [-1.3, 4.8] & $-0.3$ [-3.7, 2.8] \\
 & 1.0 & $+1.3$ [-2.7, 5.0] & $+0.8$ [-2.2, 4.2] & $+2.0$ [-1.3, 5.3] & $+1.5$ [-1.8, 4.8] & $+1.2$ [-2.3, 4.8] & $+0.5$ [-3.2, 4.2] \\
 & 1.3 & $+1.8$ [-2.8, 7.0] & $-0.3$ [-4.3, 3.5] & $+0.8$ [-3.8, 5.5] & $+0.8$ [-3.5, 5.2] & $-1.8$ [-6.5, 2.8] & $+3.7$ [-0.7, 8.7] \\
\midrule
Hermes-3-8B & 0.7 & $-1.3$ [-8.7, 6.0] & $+0.0$ [-8.7, 8.0] & $+1.3$ [-6.0, 8.0] & $+2.0$ [-4.7, 8.7] & $\mathbf{+12.0}$ [4.7, 19.3] & $+1.3$ [-6.7, 9.3] \\
 & 1.0 & $+4.7$ [-2.7, 12.0] & $-0.7$ [-8.0, 6.7] & $-0.7$ [-8.7, 7.3] & $\mathbf{+8.0}$ [1.3, 14.7] & $+4.0$ [-1.3, 9.3] & $+2.0$ [-6.7, 10.7] \\
 & 1.3 & $\mathbf{+12.7}$ [5.3, 20.0] & $\mathbf{+8.7}$ [2.0, 16.0] & $\mathbf{+20.0}$ [12.0, 29.3] & $\mathbf{+12.7}$ [4.0, 22.0] & $\mathbf{+17.3}$ [8.7, 26.7] & $\mathbf{+18.0}$ [10.0, 26.7] \\
\midrule
Qwen3.5-9B & 0.7 & $+2.0$ [-1.3, 5.3] & $-0.7$ [-5.3, 4.0] & $+1.3$ [0.0, 3.3] & $-2.0$ [-6.7, 2.0] & $-3.3$ [-8.0, 0.0] & $+0.0$ [-4.7, 4.0] \\
 & 1.0 & $+2.0$ [-1.3, 5.3] & $-2.0$ [-6.7, 1.3] & $-2.7$ [-7.3, 2.0] & $+2.7$ [-2.0, 7.3] & $+2.0$ [-4.0, 8.0] & $-1.3$ [-6.0, 3.3] \\
 & 1.3 & $\mathbf{+15.3}$ [9.3, 22.7] & $\mathbf{+15.3}$ [8.0, 24.0] & $\mathbf{+17.3}$ [10.0, 26.0] & $\mathbf{+21.3}$ [13.3, 30.0] & $\mathbf{+16.0}$ [9.3, 23.3] & $\mathbf{+19.3}$ [12.0, 28.0] \\
\midrule
OLMo-3-7B & 0.7 & $+1.3$ [-4.0, 6.7] & $-2.0$ [-8.7, 4.0] & $+1.3$ [-4.7, 8.0] & $-0.7$ [-5.3, 3.3] & $+1.3$ [-4.0, 6.7] & $+0.0$ [-6.0, 5.3] \\
 & 1.0 & $-2.0$ [-8.7, 4.7] & $-0.7$ [-8.0, 6.7] & $-0.7$ [-7.3, 6.0] & $+1.3$ [-6.0, 8.7] & $+2.7$ [-4.0, 9.3] & $+3.3$ [-4.0, 10.7] \\
 & 1.3 & $\mathbf{+10.0}$ [2.7, 17.3] & $\mathbf{+10.7}$ [4.0, 17.3] & $\mathbf{+16.0}$ [8.0, 24.0] & $\mathbf{+13.3}$ [6.0, 22.0] & $\mathbf{+18.0}$ [10.7, 26.0] & $\mathbf{+13.3}$ [6.6, 20.7] \\
\bottomrule
\end{tabular}

\end{table*}

\end{document}